\documentclass[letterpaper]{article} 
\usepackage{aaai25}  
\usepackage{times}  
\usepackage{helvet}  
\usepackage{courier}  
\usepackage[hyphens]{url}  
\usepackage{graphicx} 
\usepackage{natbib}  
\usepackage{caption} 
\usepackage{amsmath}
\usepackage{amsfonts}
\usepackage[ruled,linesnumbered]{algorithm2e}
\usepackage{subcaption}
\usepackage{multirow}
\usepackage{adjustbox}
\usepackage{booktabs}
\usepackage{arydshln}

\usepackage{newfloat}
\usepackage{listings}
\DeclareCaptionStyle{ruled}{labelfont=normalfont,labelsep=colon,strut=off} 
\title{Reinforced Multi-teacher Knowledge Distillation for Efficient General Image Forgery Detection and Localization}
\author{
    Zeqin Yu\textsuperscript{\rm 1}, Jiangqun Ni\textsuperscript{\rm 2,3}\thanks{Corresponding author.}, Jian Zhang\textsuperscript{\rm 1}, Haoyi Deng\textsuperscript{\rm 4}, Yuzhen Lin\textsuperscript{\rm 4}\\
}
\affiliations{
    \textsuperscript{\rm 1}School of Computer Science and Engineering, Sun Yat-sen University\\
    \textsuperscript{\rm 2}School of Cyber Science and Technology, Sun Yat-sen University\\
    \textsuperscript{\rm 3}Department of New Networks, Peng Cheng Laboratory\\
    \textsuperscript{\rm 4}Guangdong Key Laboratory of Intelligent Information Processing, Shenzhen University\\

}

\usepackage{bibentry}

\begin{document}

\maketitle

\begin{abstract}
Image forgery detection and localization (IFDL) is of vital importance as forged images can spread misinformation that poses potential threats to our daily lives. However, previous methods still struggled to effectively handle forged images processed with diverse forgery operations in real-world scenarios. In this paper, we propose a novel Reinforced Multi-teacher Knowledge Distillation (Re-MTKD) framework for the IFDL task, structured around an encoder-decoder \textbf{C}onvNeXt-\textbf{U}perNet along with \textbf{E}dge-Aware Module, named Cue-Net.
First, three Cue-Net models are separately trained for the three main types of image forgeries, i.e., copy-move, splicing, and inpainting, which then serve as the multi-teacher models to train the target student model with Cue-Net through self-knowledge distillation. A Reinforced Dynamic Teacher Selection (Re-DTS) strategy is developed to dynamically assign weights to the involved teacher models, which facilitates specific knowledge transfer and enables the student model to effectively learn both the common and specific natures of diverse tampering traces. Extensive experiments demonstrate that, compared with other state-of-the-art methods, the proposed method achieves superior performance on several recently emerged datasets comprised of various kinds of image forgeries.
\end{abstract}

\section{Introduction}
Recent advances in image editing and generative models~\cite{karras2019style,rombach2022high} not only enhance the quality of image manipulation and synthesis but also simplify their process. 
However, tampered images generated by such techniques could be abused to deliver misinformation, posing significant threats to social security. Accordingly, it is urgent to explore effective image forensics methods to prevent the abuse of manipulated images.

\begin{figure}[tb] 
        \centering
	\includegraphics[width=0.47\textwidth]{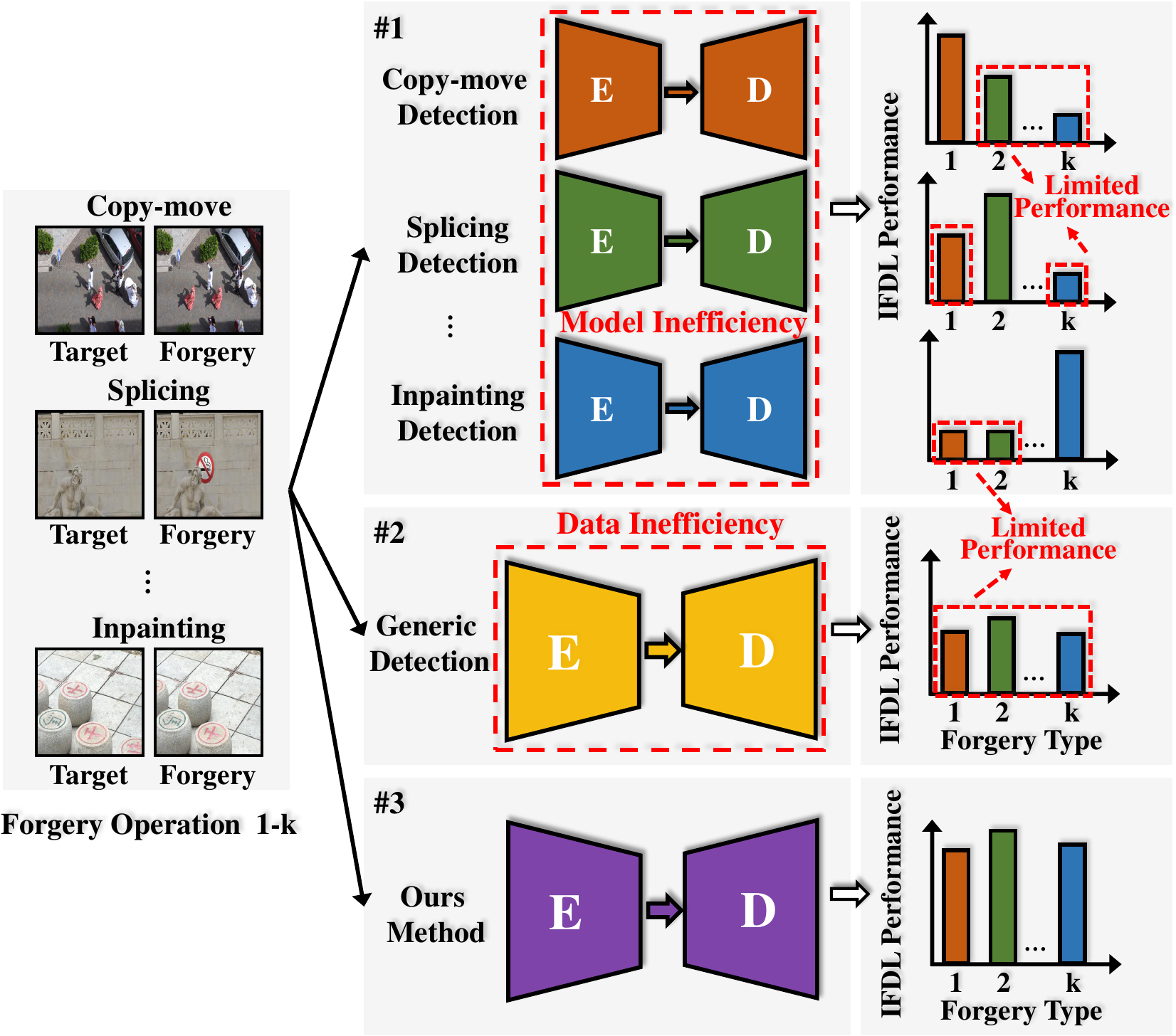}
	\caption{Overview of the existing IFDL methods. Specific IFDL methods are often limited by inefficient models, leading to poor generalization across tampering operations (the first part). Generic IFDL methods are inefficient in exploiting data and difficult to learn both the common and specific tampered features in mixed tamper data  (the second part). Our proposed method can achieve promising performance in comprehensive IFDL problems.} \label{fig:pre-task}
\end{figure}

In response to the above concern, the image forgery detection and localization (IFDL) task aims to identify forged images and localize their tampered regions.
Generally, image forgery operations mainly consist of copy-move, splicing, and inpainting.
Usually, specific methods are developed for the detection and localization of specific forgery images, which exploits the specifics of forgery trace for image tampering in a specific type,  e.g., ~\cite{wu2018busternet,chen2020serial} for copy-move, ~\cite{bi2019rru,kwon2022learning} for splicing and ~\cite{li2019localization,wu2021iid} for inpainting. 
Since the training data is limited to the specific forgery, such specific forgery detection methods may overfit and struggle with limited generalization performance. 
This limitation often results in poor performance, especially when detecting cross-source data, as illustrated in the first part of Fig.~\ref{fig:pre-task}, where the performance of an inpainting detector significantly degrades when confronted with copy-move or splicing forgeries.

To address the diverse tampering operations, generic forgery detection methods incorporate mixed data including multiple types of tampering. 
This strategy aims to capture generic tampered features
across various forgeries, e.g., noise analysis~\cite{bappy2019hybrid,zhuo2022self,guillaro2023trufor}, representation learning~\cite{hu2020span,Yu_2024_CVPR}, multi-scale supervision~\cite{dong2022mvss,ma2023imlvit} and multi-view feature fusion~\cite{liu2022pscc,guo2023hierarchical}.
However, the simultaneous learning of multiple tasks poses its own challenges, as depicted in the second part of Fig.~\ref{fig:pre-task}.
It is challenging to learn both the common and specific natures of diverse tampering traces through simple joint training alone, often leading to performance drops. 
Additionally, these methods usually entail increased complexity, computational costs, and additional parameters making it hard to meet the practical application scenarios.
Therefore, it is crucial to promote a model’s ability to discern common traits and specifics among various tampering traces and improve its generalization performance in advancing image forgery detection and localization capabilities.

In this paper, we present a novel Reinforced Multi-teacher Knowledge Distillation (Re-MTKD) framework for the IFDL task. First, we propose the Cue-Net model, built upon a \textbf{C}onvNeXt-\textbf{U}perNet structure and equipped with a novel \textbf{E}dge-Aware Module (EAM). The integration of EAM facilitates forgery artifact detection by fusing low-level and high-level features. To effectively train Cue-Net, we introduce a novel knowledge distillation strategy that utilizes multiple teacher models and activates them during training through a reinforcement learning strategy, named Reinforced Dynamic Teacher Selection (Re-DTS).
Specifically, we first train multiple teacher models on different datasets, each focused on a specific type of forgery. By incorporating the Re-DTS strategy, these well-trained teacher models are dynamically selected based on the type of tampering data for knowledge distillation. 
This approach allows the knowledge from teacher models pretrained on different forgery types to be efficiently transferred to a student model, reducing the risk of overfitting to any specific type of forgery trace.
Our contributions to this work are as follows:

\begin{itemize}
	\item We propose a Reinforced Multi-teacher Knowledge Distillation (Re-MTKD) framework, which consists of ConvNeXt-UPerNet structure with a novel Edge-Aware Module (EAM), which fuses low-level and high-level features, significantly enhancing the ability to detect tampering traces.
	
	\item We propose a Reinforced Dynamic Teacher Selection (Re-DTS) strategy to maximize the forgery detection and localization capability of the student model by dynamically selecting teacher models that excel in specific tamper forensic tasks for dedicated knowledge transfer, enabling the student model to learn the commonalities and specifics of multiple tamper traces.
	
	\item Comprehensive experiments show that the proposed method achieves superior performances on several recently multiple tampering types of datasets compared with other state-of-the-art methods.

\end{itemize}

\section{Related Work}
\subsection{Image Forgery Detection and Localization}
There are several algorithms for IFDL tasks, including specific forgery detection for copy-move, splicing, inpainting, and generic forgery detection strategies.

\noindent \textbf{Specific forgery detection.} 
Specific IFDL methods focus on identifying specific tampering traces for specific types of tampered data.
For \textit{copy-move detection}, 
Buster-Net~\cite{wu2018busternet} and CMSD-STRD~\cite{chen2020serial} were designed to localize source/target regions in copy-move tampered images in parallel and in series, respectively.
For \textit{splicing detection}, MFCN~\cite{salloum2018image} focuses on highlighting tampered edges, RRU-Net~\cite{bi2019rru} emphasizes residual artifacts, and CAT-Net~\cite{kwon2022learning} analyzes compression artifacts in RGB and DCT domains.
For \textit{inpainting detection}, HP-FCN~\cite{li2019localization} uses high-pass filtering to enhance inpainting traces, IID-Net~\cite{wu2021iid} employs neural architecture search for automatic feature extraction, and TLTF-LEFF~\cite{li2023transformer} introduces a local enhancement transformer architecture.
While these methods achieve excellent performance on specific forgery types, they often struggle with other types of tampered data, leading to inefficient model generalization.

\noindent \textbf{Generic forgery detection.} 
With the growth of massive data, recent generic IFDL methods aim to learn the commonalities of various tampering tasks.
H-LSTM~\cite{bappy2019hybrid}, SATL-Net~\cite{zhuo2022self}, CFL-Net~\cite{niloy2023cfl} and TruFor~\cite{guillaro2023trufor} integrate spatial and noise domains to detect diverse manipulations. SPAN~\cite{hu2020span} and DiffForensics~\cite{Yu_2024_CVPR} leverage self-supervised learning to identify subtle tampering traces.
MVSS-Net~\cite{dong2022mvss} and IML-Vit~\cite{ma2023imlvit} emphasize multi-scale supervision, while PSCC-Net~\cite{liu2022pscc} and HiFi-Net~\cite{guo2023hierarchical} focus on multi-view feature fusion.
Although these methods achieve promising results across various forgery operations, they often suffer performance degradation in joint training due to task incompatibilities and often introduce extra complexity and computational cost.

\begin{figure*}[tb] 
        \centering
	\includegraphics[width=0.99\textwidth]{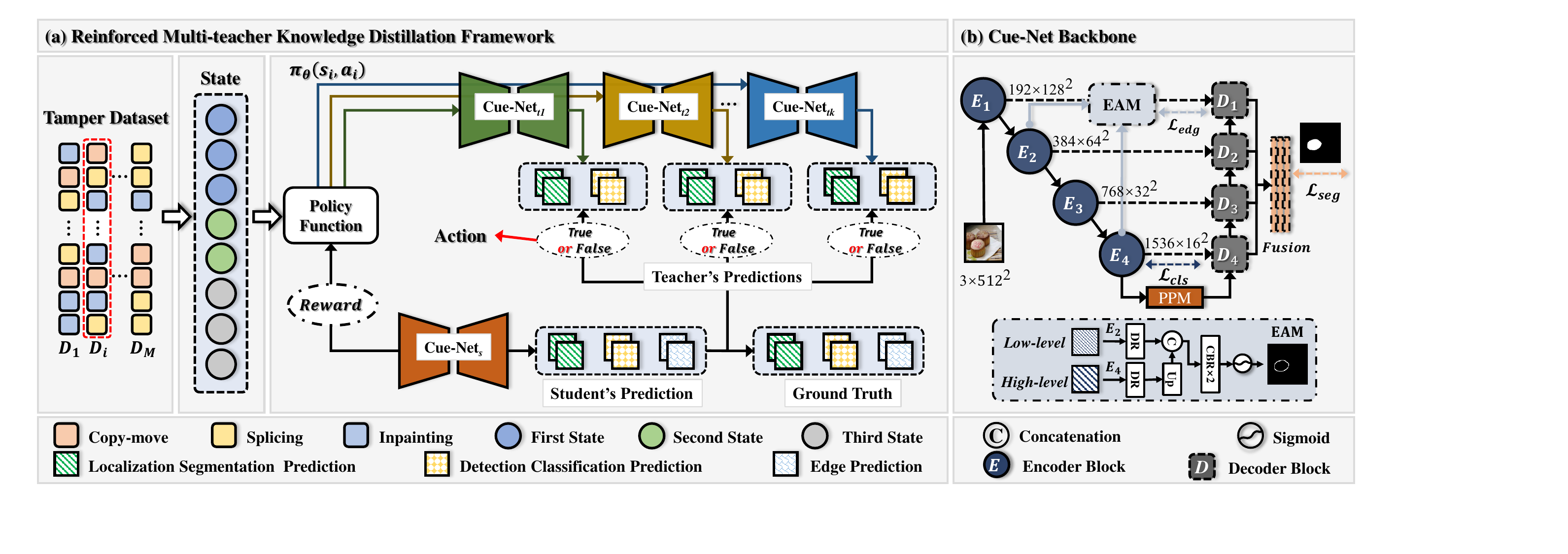}
	\caption{We propose a novel Reinforced Multi-teacher Knowledge Distillation framework, structured by the simple yet effective Cue-Net backbone, for the IFDL task. Within this framework,
 the proposed Re-DTS strategy dynamically selects teacher models based on different tampering types of data, guiding the student model to effectively learn various tampering traces.
 } 
 \label{fig:main_frame}
\end{figure*}

\subsection{Knowledge Distillation}
Initial knowledge distillation (KD)~\cite{hinton2015distilling} transfers knowledge from a large teacher model to a smaller student model via a teacher-student architecture.
This concept was extended by~\cite{Zhang_2019_ICCV} into self-distillation, where the same network serves as both teacher and student, allowing the student to learn from itself.
For IFDL tasks,
\cite{yu2023learning} proposes a fixed-weight multi-teacher self-distillation strategy to handle five tampering operations in smartphone screenshot images, enabling the student model to learn multiple tampering traces.
However, assigning fixed weights to each teacher model may not be optimal, as it prevents dynamic adaptation to different data batches, which can contain varied tampering types, potentially limiting the IFDL performance of the student model.
\subsection{Reinforcement Learning}
Reinforcement learning has shown promising results in areas such as autonomous driving~\cite{sinha2020d2rl}, smart games~\cite{silver2016mastering}, and recommendation systems~\cite{feng2018reinforcement,yuan2021reinforced}. Its core idea is to maximize long-term rewards by enabling an agent to learn from interactions with the environment and adjust its behavior based on reward signals. In this paper, we introduce a novel Re-MTKD framework that incorporates these principles. Central to this framework is the Re-DTS strategy, dynamically selecting the most appropriate teacher models to transfer specialized knowledge to the student model. This strategy enhances the student model's ability to handle various tampering traces and improves IFDL performance.

\section{Proposed Re-MTKD Framework}
\subsection{Cue-Net Backbone}
\noindent \textbf{Pipeline.} 
The architecture of Cue-Net is shown in Fig.~\ref{fig:main_frame}(b), which is an encoder-decoder structure.
For the encoder, assuming the input image is $x \in \mathbb{R}^{h \times w \times c} $, feature extraction is first performed by ConvNeXt v2~\cite{woo2023convnext} and feature maps $\left \{ E_{j} \right \} _{j=1}^{4} $ are obtained from low to high levels.
Additionally, $E_{2}$ and $E_{4}$ are used as inputs to the EAM proposed in this paper, which will be presented in the next subsection.
The decoder employs UPerNet~\cite{xiao2018unified}, comprising two components: the Pyramid Pooling Module (PPM)~\cite{zhao2017pyramid} and the Feature Pyramid Network (FPN)~\cite{lin2017feature}. The FPN outputs feature maps $\left \{ D_{j} \right \} _{j=1}^{4} $, which are resized to their original dimensions using bilinear interpolation and then merged to obtain the localization segmentation result after applying the sigmoid activation function.

\noindent \textbf{Edge-Aware Module.}
To further enhance fine-grained tampered trace extraction, we design a simple yet effective edge-aware module (EAM) that accurately extracts the edge information of tampered regions for subsequent guidance and fusion.
Tampered region edges usually consist of only a few pixels, making it difficult to achieve accurate edge supervision using only high-level feature maps with the lowest resolution. Conversely, using only low-level feature maps with high resolution makes it hard to sense the tampered area.
In EAM, we perform edge supervision on multi-level features, which is derived by  fusing low-level feature $E_{2}$ and high-level feature $E_{4}$ in the encoder, shown as follows,
\begin{equation}
    f^{e} = \sigma (CBR(Cat(DR(E_{2}), Up(DR(E_{4}))))),
\end{equation}
where $\sigma(\cdot)$ is sigmoid normalization and the CBR blocks (Conv + BN + ReLU) implement the fusion of low-level and high-level features, $Cat(\cdot)$ forms a concatenation operation, $DR$ stands for dimension reduction using the CBR blocks and $Up$ for upsampling.
Later, the obtained edge prediction map $f^{e}$ and edge label $y^{e}$ are used for loss iteration.

\noindent \textbf{Loss function.}
Following the previous IFDL methods \cite{liu2022pscc, dong2022mvss,guo2023hierarchical}, our method includes three types of supervision: localization segmentation (pixel-level) supervision $\mathcal{L}_{seg}$ and detection classification (image-level) supervision $\mathcal{L}_{cls}$ as standard, and edge detection supervision $\mathcal{L}_{edg}$ as a special design.

For localization segmentation supervision $\mathcal{L}_{seg}$, following \cite{guillaro2023trufor}, we use a combination of weighted $\ell _{wbce}$ and $\ell _{dice}$~\cite{milletari2016v}.
\begin{equation}
	\mathcal{L}_{seg} =\lambda _{0}^{s}\ell _{wbce}(f^{s}, y^{s})  +\left ( 1-\lambda _{0}^{s}  \right )\ell _{dice}(f^{s}, y^{s}), 
\end{equation}
where $\lambda _{0}^{s}$ is the segmentation balance weight, 
$f^{s}$ and $y^{s}$ are the localization results and localization labels, respectively.

Following ConvNeXt v2~\cite{woo2023convnext}, we design the detection classification loss $\mathcal{L}_{cls}$ on $E_{4}$ using standard $\ell_{bce}$,
\begin{equation}
	\mathcal{L}_{cls}= \ell_{bce}(f^{c}, y^{c}),
\end{equation}
where $f^{c}$ and $y^{c}$ are the detection classification results and classification labels, respectively.

For edge supervision $\mathcal{L}_{edg}$ with EAM, we use $\ell _{dice}$ to better focus on tiny edge regions, 
\begin{equation}
	\mathcal{L}_{edg}= \ell _{dice}(f^{e}, y^{e}).
\end{equation}

For the final combined loss, which is also the ``hard'' loss $\mathcal{L}_{\textbf{hard}}$, weighting from three perspectives yields our nearest supervised loss function,
\begin{equation}
	\mathcal{L}_{\textbf{hard}} = \alpha \cdot \mathcal{L}_{seg} + \beta \cdot (\mathcal{L}_{cls} + \mathcal{L}_{edg}),
	\label{all loss}
\end{equation}
where $\alpha ,\beta \in \left [ 0,1 \right ] $. The term ``hard'' indicates
the binary label for each pixel is used for supervision.

\subsection{Reinforced Dynamic Teacher Selection Strategy }

Before introducing the Re-DTS strategy, we will first formulate single-teacher KD. To better improve both the performance of localization segmentation and detection classification of the student model, we choose to transfer specialized knowledge through ``soft'' loss $\mathcal{L}_{\textbf{soft}}$ shown as follows,
\begin{equation}
    \mathcal{L}_{\textbf{soft} } =  \mathcal{L}_{seg}(f_{\textbf{s}}^{s}, f_{\textbf{t}}^{s}) + \mathcal{L}_{cls}(f_{\textbf{s}}^{c}, f_{\textbf{t}}^{c}),
    \label{Eq:single-teacher}
\end{equation}
where bolded \textbf{t} and  \textbf{s} denote the teacher model and student model, respectively.

Figure~\ref{fig:main_frame}(a) shows the overview of our Re-DTS strategy.
The elements (i.e., \textit{state}, \textit{action}, and \textit{reward}) of reinforcement learning are introduced as follows. To be clear, since all the teacher models share the same training process, we will introduce only one of them and the three teacher models are denoted as $\left \{ \Theta _{k}^{t} \right \} _{k=1}^{3} $.

\noindent \textbf{State.} 
Our Re-DTS strategy contains a series of environment states, $s_{1}$, $s_{2}$, ..., which summarise the features of the tampered image to be detected, student model and the candidate teacher model so that reasonable decisions can be made. To formulate the state $s_{i}$, we design a state feature vector \textbf{F}$(s_{i})$ containing a concatenation of three features.

The first feature is the vector representation $\mathcal{R}(x_{i}) \in \mathbb{R}^{d} $ of $i$-th batch image $x_{i}$, which could be generated by any IFDL representation model $\Theta ^{r}$. In this paper, we obtain the output of the last layer of the decoder $D_{4}$ as the feature $\mathcal{R}(x_{i})$ by feeding this batch image $x_{i}$ to Cue-Net, which has not been trained on any IFDL task data.

The second feature $\mathcal{S}(x_{i}) \in \mathbb{R}^{d}$ is obtained similarly, except the feature vector is output from the student model $\Theta^{s}$, trained in real-time on the IFDL task data and designed to effectively represent the content of the tampered image.

The third feature $\mathcal{T}_{k}(x_{i}) \in \mathbb{R}^{1+1+1}$, consisting of the localization segmentation prediction result $f_{\textbf{t}k}^{s}$, the detection classification prediction result $f_{\textbf{t}k}^{c}$ and the tampering edge prediction result $f_{\textbf{t}k}^{e}$ of the teacher model $\Theta _{k}^{t}$, respectively, are expected to guide the better action execution of the policy network.
\begin{equation}
\mathcal{T}_{k}(x_{i} ) = Cat(Max(f_{\textbf{t}k}^{s}), f_{\textbf{t}k}^{c}, Max(f_{\textbf{t}k}^{e})),
\end{equation}
where $Max(\cdot)$ is the maxpooling operation.

Therefore, the state feature vector $\textbf{F}\left ( s_{i} \right )\in \mathbb{R}^{2d+3}$ can be formulated as follows:
\begin{equation}
\textbf{F}\left ( s_{i} \right ) = Cat(\mathcal{R}(x_{i}), \mathcal{S}(x_{i}), \mathcal{T}_{k}(x_{i})),
\end{equation}
where such concatenation ensures that the state feature vector $\textbf{F}\left ( s_{i} \right )$ comprehensively reflects the image and model characteristics, enabling more informed and effective decisions by the policy network.

\noindent \textbf{Action.}
Each teacher model $\Theta ^{t}$ is associated with a policy network, and actions $a_{i} \in \left \{ 0,1 \right \} $ are defined to indicate whether the policy network will be chosen to transfer the specialized knowledge of the $k$-th teacher model $\Theta _{k}^{t}$ to the student model $\Theta ^{s}$.
We sample the value of $a_{i}$ using the policy function $\pi _{\theta } \left ( s_{i}, a_{i}  \right ) $, where $\theta$ is the parameter to be learned. In this paper, we adopt a logistic function as the policy function:
\begin{equation}
	\begin{split}
        \pi_\theta\left ( s_{i}, a_{i}  \right ) &= P_\theta \left ( a_{i}\mid s_{i}\right ) \\ 
        &= a_{i} \sigma (\textbf{W} \ast \textbf{F}\left ( s_{i} \right )+b  ) \\
        &+ \left ( 1-a_{i} \right )\left ( 1- \sigma (\textbf{W} \ast \textbf{F}\left ( s_{i} \right )+b  )\right ),
	\end{split}
\end{equation}
where $\textbf{W} \in \mathbb{R}^{2d+3}$ and $ b \in \mathbb{R}^{1}$ are the trainable parameters.

\noindent \textbf{Reward.}
Reasonable rewards can instruct the teacher model to distill a high-performing student model.
For training data $\mathcal{D} = \left \{ \mathcal{D}_{1}, \mathcal{D}_{2}, ...,\mathcal{D}_{M} \right \}$, with $M$ is the number of batches per epoch, we construct a feature vector $s_{ik}$ for each teacher model $\Theta _{k}^{t}$ and sample action $a_{ik}$ for the $i$-th batch of data $\mathcal{D}_{i}$ according to the policy function $\pi _{\theta } \left ( s_{ik}, a_{ik}  \right ) $ to determine whether the action is selected or not.
For all sampled teacher models, we integrate the loss $\mathcal{L}_{\textbf{soft}}$ of all teacher models with the loss $\mathcal{L}_{\textbf{hard}}$ to update the student model parameters $\Theta ^{s}$.

\begin{algorithm}[tb]
\caption{Training Procedure for Re-MTKD}
\label{alg:alogorithm1}
    Pre-train teacher models $\Theta _{k}^{t}$ using the corresponding 
    type of tampered data $\mathcal{D}^{k}$.\\
    Pre-train the policy network $\theta _{k}$ by calculating the return
    under all teacher models $\Theta _{k}^{t}$ selected. \\
    Run Algorithm~\ref{alg:alogorithm2} to jointly train the student model $\Theta ^{s}$ and the policy network $\theta _{k}$ until convergence.
\end{algorithm}

\begin{algorithm}[tb]
    \caption{Re-DTS strategy}
    \label{alg:alogorithm2}
    \KwIn{Epoch number $L$. Batch-size number $B$. Training data
    $\mathcal{D} = \left \{ \mathcal{D}_{1}, \mathcal{D}_{2}, ...,\mathcal{D}_{M} \right \}$. $k$-th teacher model and policy network initialized as $\Theta ^{t} = \Theta _{k}^{t}$ and $\theta = \theta _{k}$.}
    \For{$epoch$ $l = 1$ to $L$}{
    Shuffle $\mathcal{D}$ to obtain a new training sequence.\\
    \For{each batch $\mathcal{D}_{i} \in \mathcal{D}$}{
    Sample teacher selection actions for each batch image $x_{i} \in \mathcal{D}_{i}$ with $\theta$ by: $a_{i} \sim \pi_\theta\left ( s_{i}, a_{i}  \right )$. \\
    Stored $\left (a_{i}, s_{i} \right )$ to the episode history $\mathcal{H}$.\\
    Compute the soft labels of the selected teachers $\mathcal{K}$:
    $\mathcal{L}_\textbf{soft} = {\textstyle \sum_{k\in\mathcal{K}}}\mathcal{L}_{\textbf{soft}}^{(k)}$ \\
    Update the parameter $\Theta ^{s}$ of student model by:
    $\mathcal{L} = \mathcal{L}_{\textbf{hard}} + \omega\mathcal{L}_{\textbf{soft}}$ \\
    \If{$i$ mod $10B$}{
    \For{each $\left (a_{i}, s_{i} \right ) \in \mathcal{H} $}{
    Compute delayed reward following Eq.~\ref{Eq:reward}.\\
    Update the parameter $\theta$ of the policy function following Eq.~\ref{Eqa:Optimization}.
    }
    }
    }
    }
\end{algorithm}

\begin{table*}[]
        \setlength\tabcolsep{1.4mm}
	\centering
        \small
	\begin{tabular}{clccc|ccc|ccc|ccc|ccc}
		\hline
		\multicolumn{2}{c}{\multirow{2}{*}{\textbf{Methods}}} & \multicolumn{3}{c}{\textbf{Copy-move}} & \multicolumn{3}{c}{\textbf{Splicing}} & \multicolumn{3}{c}{\textbf{Inpainting}} & \multicolumn{3}{c}{\textbf{Multi}} & \multicolumn{3}{c}{\textbf{Average}} \\ 
        \cline{3-17} 
		\multicolumn{2}{c}{} & Acc & F1 & AUC & Acc & F1 & AUC & Acc & F1 & AUC & Acc & F1 & AUC & Acc & F1 & AUC \\ 
        \hline
		\multirow{3}{*}{\rotatebox{90}{\centering D-Com}} & DoaGan$^{*}$ & 0.695 & 0.730 & 0.756 & 0.463 & 0.432 & 0.516 & 0.383 & 0.351 & 0.522 & 0.487 & 0.406 & 0.560 & 0.507 & 0.479 & 0.588 \\
		& Buster-Net$^{*}$ & 0.557 & 0.715 & 0.500 & 0.638 & 0.779 & 0.500 & 0.726 & 0.841 & 0.500 & 0.613 & 0.760 & 0.500 & 0.634 & 0.774 & 0.500 \\
		& CMSD-STRD & 0.472 & 0.332 & 0.502 & 0.475 & 0.445 & 0.530 & 0.517 & 0.585 & 0.558 & 0.472 & 0.478 & 0.494 & 0.484 & 0.460 & 0.521 \\ \hline
		\multirow{3}{*}{\rotatebox{90}{\centering D-Spl}} & MFCN & 0.529 & 0.692 & 0.422 & 0.637 & 0.779 & 0.586 & 0.725 & 0.841 & 0.615 & 0.621 & \underline{0.764} & 0.565 & 0.628 & 0.769 & 0.547 \\
		& RRU-Net & \underline{0.828} & \underline{0.865} & \underline{0.863} & 0.818 & 0.873 & 0.872 & 0.824 & 0.890 & 0.899 & 0.612 & 0.756 & 0.624 & \underline{0.770} & \underline{0.846} & 0.815 \\
		& CAT-Net & 0.719 & 0.758 & 0.777 & 0.805 & 0.861 & 0.917 & 0.831 & 0.888 & 0.887 & 0.611 & 0.730 & 0.632 & 0.742 & 0.809 & 0.803 \\ \hline
		\multirow{3}{*}{\rotatebox{90}{\centering D-Inp}} & HP-FCN & 0.557 & 0.715 & 0.540 & 0.636 & 0.777 & 0.566 & 0.726 & 0.841 & 0.571 & 0.613 & 0.760 & \underline{0.703} & 0.633 & 0.774 & 0.595 \\
		& IID-Net & 0.505 & 0.298 & 0.809 & 0.648 & 0.642 & 0.888 & 0.670 & 0.748 & 0.748 & 0.526 & 0.473 & 0.639 & 0.587 & 0.540 & 0.771 \\
        & TLTF-LEFF & 0.446 & 0.010 & 0.755 & 0.386 & 0.078 & 0.808 & 0.474 & 0.433 & 0.790 & 0.392 & 0.025 & 0.646 & 0.425 & 0.136 & 0.750 \\\hline
		\multirow{6}{*}{\rotatebox{90}{\centering D-Generic}} & H-LSTM & 0.557 & 0.715 & 0.501 & 0.638 & 0.779 & 0.516 & 0.726 & 0.841 & 0.517 & 0.613 & 0.760 & 0.507 & 0.634 & 0.774 & 0.510 \\
		& SPAN &0.557 & 0.715 & 0.500 & 0.638 & 0.779 & 0.500 & 0.726 & 0.841 & 0.500 & 0.613 & 0.760 & 0.500 & 0.634 & 0.774 & 0.500 \\
		& MVSS-Net & 0.598 & 0.494 & 0.630 & 0.843 & 0.862 & 0.872 & 0.835 & 0.882 & 0.833 & \underline{0.641} & 0.638 & 0.681 & 0.729 & 0.719 & 0.754 \\
		& SATL-Net & 0.713 & 0.717 & 0.815 & 0.854 & 0.881 & \underline{0.934} & \underline{0.867} & \underline{0.909} & \underline{0.920} & 0.634 & 0.694 & 0.666 & 0.767 & 0.800 & \underline{0.834} \\
		& PSCC-Net & 0.473 & 0.614 & 0.467 & 0.657 & 0.785 & 0.721 & 0.706 & 0.827 & 0.537 & 0.615 & 0.761 & 0.610 & 0.613 & 0.747 & 0.584 \\ 
		& HiFi-Net & 0.557 & 0.715 & 0.500 & 0.638 & 0.779 & 0.500 & 0.726 & 0.841 & 0.500 & 0.613 & 0.760 & 0.500 & 0.634 & 0.774 & 0.500 \\ 
  		& IML-Vit & 0.675 & 0.657 & 0.817 & \underline{0.869} & \underline{0.899} & 0.933 & 0.774 & 0.837 & 0.831 & 0.639 & 0.707 & 0.681 & 0.739 & 0.775 & 0.256 \\ 
            \hline
		\multicolumn{2}{c}{Re-MTKD (Ours)} & \textbf{0.888} & \textbf{0.900} & \textbf{0.952} & \textbf{0.893} & \textbf{0.918} & \textbf{0.970} & \textbf{0.973} & \textbf{0.981} & \textbf{0.994} & \textbf{0.703} & \textbf{0.778} & \textbf{0.790} & \textbf{0.864} & \textbf{0.894} & \textbf{0.927} \\ 
            \hline
	\end{tabular}
     \caption{Image-level Acc, F1 under 0.5 threshold, and AUC scores for image forgery detection.
     The \textbf{best} and \underline{2nd-best} results are highlighted. 
     Methods with $^{*}$ use the original paper's pre-trained model, others keep the same training data as our method.
     }
 \label{tab:detection result}
\end{table*}

We propose three reward calculations. The first is the ``hard'' loss $\mathcal{L}_{\textbf{hard}}$, comprising localization segmentation loss $\mathcal{L}_{seg}$, detection classification loss $\mathcal{L}_{cls}$ and tampering edge detection loss $\mathcal{L}_{edg}$.
The second reward function combines the ``hard'' loss $\mathcal{L}_{\textbf{hard}}$ as well as the ``soft'' loss $\mathcal{L}_{\textbf{soft}}$ of the teacher model.
Finally, to enhance the performance of the student model on the IFDL task, we use the localization segmentation F1 score and detection classification accuracy score of the student model during training as the third reward.
In summary, we have
 \begin{equation}
	\begin{split}
        reward_{1} & = -\mathcal{L}_{\textbf{hard}} \\
        reward_{2} & = -(\mathcal{L}_{\textbf{hard}} + \mathcal{L}_{\textbf{soft}}) \\
        reward_{3} & = -\gamma(\mathcal{L}_{\textbf{hard}} + \mathcal{L}_{\textbf{soft}}) \\
        & + (1-\gamma)((F1_{seg} + Acc_{cls}) on \mathcal{D}_{i}),
	\end{split}
 \label{Eq:reward}
\end{equation}
where $\gamma$ is a hyperparameter to balance the different rewards. 
We ultimately take the third reward as the default option.
Notably, following REINFORCE~\cite{williams1992simple}, where updates are not rewarded immediately after taking each step, we delayed updates until after $10B$ batches of training, where $B$ notes the batch size.

\noindent \textbf{Optimization.} 
According to the policy gradient theorem~\cite{sutton1999policy} and the REINFORCE algorithm~\cite{williams1992simple}, we compute the gradient to update the current policy as follows:
\begin{equation}
    \theta \gets  \theta + \xi \sum_{i}^{} r\sum_{k}^{} \nabla _{\theta } \pi _{\theta }(s_{ik},a_{ik}),
    \label{Eqa:Optimization}
\end{equation}
where $r$ is the reward function defined in Eq.~\ref{Eq:reward}, and $\xi$ is the learning rate.

\begin{table*}[tb]
        \setlength\tabcolsep{1.4mm}
        \small
	\centering
        \begin{tabular}{clccc|ccc|ccc|ccc|ccc}
		\hline
		\multicolumn{2}{c}{\multirow{2}{*}{\textbf{Methods}}} & \multicolumn{3}{c}{\textbf{Copy-move}} & \multicolumn{3}{c}{\textbf{Splicing}} & \multicolumn{3}{c}{\textbf{Inpainting}} & \multicolumn{3}{c}{\textbf{Multi}} & \multicolumn{3}{c}{\textbf{Average}} \\ 
        \cline{3-5} \cline{6-17}
		\multicolumn{2}{c}{} & F1 & IoU & AUC & F1 & IoU & AUC & F1 & IoU & AUC & F1 & IoU & AUC & F1 & IoU & AUC \\ \hline
		\multirow{3}{*}{\rotatebox{90}{\centering D-Com}} & DoaGan$^{*}$ & 0.249 & 0.169 &  \underline{0.785} & 0.027 & 0.021 & 0.550 & 0.006 & 0.004 & 0.561 & 0.020 & 0.013 & 0.569 & 0.076 & 0.052 & 0.616 \\
		& Buster-Net$^{*}$ & \underline{0.274} & 0.185 & \textbf{0.806} & 0.124 & 0.083 & 0.715 & 0.048 & 0.028 & 0.666 & 0.077 & 0.048 & 0.705 & 0.131 & 0.086 & 0.723 \\
		& CMSD-STRD & 0.018 & 0.011 & 0.507 & 0.037 & 0.023 & 0.505 & 0.051 & 0.032 & 0.514 & 0.018 & 0.011 & 0.504 & 0.031 & 0.019 & 0.508 \\ \hline
		\multirow{3}{*}{\rotatebox{90}{\centering D-Spl}} & MFCN & 0.141 & 0.082 & 0.634 & 0.256 & 0.161 & 0.707 & 0.440 & 0.327 & 0.735 & 0.154 & 0.091 & 0.669 & 0.248 & 0.165 & 0.686 \\
		& RRU-Net & 0.165 & 0.113 & 0.618 & 0.454 & 0.378 & 0.806 & \underline{0.604} & \underline{0.521} & \underline{0.885} & 0.232 & 0.178 & 0.687 & 0.364 & 0.298 & 0.749 \\
		& CAT-Net & 0.270 & \textbf{0.224} & 0.746 & \underline{0.622} & \underline{0.560} & 0.877 & 0.429 & 0.345 & 0.817 & 0.240 & 0.190 & 0.717 & \underline{0.390} & \underline{0.330} & 0.789 \\ \hline
		\multirow{3}{*}{\rotatebox{90}{\centering D-Inp}} & HP-FCN & 0.012 & 0.007 & 0.685 & 0.013 & 0.007 & 0.699 & 0.008 & 0.004 & 0.713 & 0.058 & 0.033 & 0.728 & 0.023 & 0.013 & 0.706 \\
		& IID-Net & 0.022 & 0.016 & 0.648 & 0.157 & 0.124 & 0.795 & 0.273 & 0.214 & 0.858 & 0.043 & 0.031 & 0.703 & 0.124 & 0.096 & 0.751 \\
		& TLTF-LEFF & 0.000 & 0.000 & 0.615 & 0.010 & 0.007 & 0.713 & 0.157 & 0.132 & 0.723 & 0.037 & 0.003 & 0.711 & 0.043 & 0.035 & 0.691 \\ \hline
		\multirow{6}{*}{\rotatebox{90}{\centering D-Generic}} & H-LSTM & 0.143 & 0.081 & 0.602 & 0.199 & 0.120 & 0.617 & 0.344 & 0.229 & 0.656 & 0.134 & 0.078 & 0.604 & 0.205 & 0.127 & 0.620 \\
		& SPAN & 0.052 & 0.032 & 0.506 & 0.098 & 0.063 & 0.681 & 0.455 & 0.350 & 0.736 & 0.137 & 0.092 & 0.677 & 0.186 & 0.134 & 0.650 \\
		& MVSS-Net & 0.165 & 0.137 & 0.584 & 0.521 & 0.464 & 0.759 & 0.571 & 0.488 & 0.769 & 0.241 & \underline{0.198} & 0.630 & 0.374 & 0.322 & 0.686 \\
		& SATL-Net & 0.073 & 0.050 & 0.606 & 0.284 & 0.221 & 0.738 & 0.328 & 0.241 & 0.804 & 0.078 & 0.054 & 0.616 & 0.191 & 0.142 & 0.691 \\
		& PSCC-Net & 0.105 & 0.066 & 0.612 & 0.330 & 0.241 & 0.745 & 0.368 & 0.274 & 0.579 & \underline{0.278} & 0.194 & 0.766 & 0.270 & 0.194 & 0.676 \\ 
		& HiFi-Net & 0.161 & 0.093 & 0.642 & 0.233 & 0.145 & 0.599 & 0.210 & 0.138 & 0.587 & 0.139 & 0.084 & 0.577 & 0.186 & 0.115 & 0.601 \\ 
  	& IML-Vit & 0.160 & 0.136 & 0.773 & 0.498 & 0.446 & \underline{0.903} & 0.215 & 0.166 & 0.795 & 0.130 & 0.100 & \underline{0.772} & 0.251 & 0.212 & \underline{0.810} \\  
         \hline
		\multicolumn{2}{c}{Re-MTKD (Ours)} & \textbf{0.277} & \underline{0.221} & 0.764 & \textbf{0.676} & \textbf{0.606} & \textbf{0.934} & \textbf{0.789} & \textbf{0.714} & \textbf{0.965} & \textbf{0.444} & \textbf{0.372} & \textbf{0.858} & \textbf{0.531} & \textbf{0.468} & \textbf{0.861} \\ \hline
	\end{tabular}
         \caption{Pixel-level F1, IoU under 0.5 threshold, and AUC scores for image forgery localization. Same highlighting conventions and conditions as in Table~\ref{tab:detection result} apply.}
    \label{tab:localization result}
\end{table*}
\subsection{Model Training}
Algorithm~\ref{alg:alogorithm1} outlines the training procedure for the Re-MTKD framework.
We pre-train the model before starting joint training. 
Teacher models $\Theta_{k}^{t}$ are pre-trained with specialized knowledge on the corresponding type of tampered data $\mathcal{D}^{k}$. Then, the policy network $\theta_{k}$ is initialized by selecting feedback from all teacher models during the KD process.

After initialization, the Re-DTS strategy is executed to optimize the student model according to $\mathcal{L}$, as shown in Algorithm~\ref{alg:alogorithm2}.
In this process, we first fix the policy network $\theta_{k}$ to train the student models $\Theta^{s}$. After a pre-determined number of iterations, we fix $\Theta^{s}$, compute the return content, and optimize the teacher selection policy network $\theta_{k}$. This process is repeated until all training iterations are completed.

\section{Experiments}
\subsection{Experimental Setup}
\noindent \textbf{Dataset.}
Considering the availability and generality, we select ten challenging benchmark datasets to evaluate our method, covering tampering types: copy-move (Com), splicing (Spl), inpainting (Inp), and multi-tampering (Multi).
Details of these datasets are provided in the Appendix.

1) For Com,
we use CASIA v2~\cite{dong2013casia} and Tampered Coco~\cite{liu2022pscc} for training and CASIA v1+~\cite{dong2013casia}, NIST16~\cite{nimble2016datasets} and Coverage~\cite{wen2016coverage} for testing.

2) For Spl, we use CASIA v2~\cite{dong2013casia} and Fantastic-Reality~\cite{kniaz2019point} for training, and CASIA v1+~\cite{dong2013casia}, NIST16~\cite{nimble2016datasets}, Columbia~\cite{hsu2006detecting} and DSO-1~\cite{de2013exposing} for testing.

3) For Inp,
we use GC Dresden\&Places~\cite{wu2021iid} for training, and NIST16~\cite{nimble2016datasets}, AutoSplicing~\cite{jia2023autosplice} and DiverseInp~\cite{wu2021iid} for testing.

4) For Multi, we use IFC~\cite{IFC}, Korus~\cite{de2013exposing}, and IMD2020~\cite{novozamsky2020imd2020} for testing, as this data involves composite operations of the single tampering types described above.

For the Cue-Net student model $\Theta^{s}$ and the other SOTA methods, training is conducted on all mixed single-tampered data. In contrast, during reinforced multi-teacher knowledge distillation, each teacher model $\Theta^{t}$ is trained on the corresponding type of tampered data.

\noindent \textbf{Implementation details.}
We use a single Nvidia Tesla A100 GPU (80 GB memory) to conduct experiments on the PyTorch deep learning framework. The parameter configurations for reinforced multi-teacher KD are as follows:

1) For single-teacher model pre-training,
we resize the input image to 512$\times$512 and apply the AdamW optimizer. 
We set the training hyper-parameters by the learning rate as $1\times10^{-4}$, the batch size as 24, and the epoch as 50. 
To balance the performance of forgery detection and localization, we set the weight of forgery localization $\mathcal{L} _{seg}$ to $\alpha=1$ and $\lambda _{0}^{s}$ is 0.1.
The weight $\beta$ of the detection classification supervision $\mathcal{L} _{cls}$ and edge supervision $\mathcal{L} _{edg}$ is set to 0.2.

2) For reinforced multi-teacher KD,
we use the same parameter settings for input image size, optimizer, learning rate, and batch size as for the single-teacher model pre-training but with more training data and only 25 epochs at this training stage.
For $\gamma$ in $reward_{3}$, we set it to 0.2.
For $\mathcal{L}$, we set the constraint factor $\omega=0.05$ for the overall teacher loss $\mathcal{L}_{\textbf{soft}}$, and this constraint factor is multiplied by the number of selected teachers to balance the process of teacher model knowledge transfer and student model self-learning.

3) For the policy network,
the learning rate is set to $3\times10^{-4}$ and adjusted using CosineAnnealingLR with the Adam optimizer.

\subsection{Comparison with the State-of-the-Art Methods}
For a fair comparison, we focus on methods with available codes or pre-trained models trained on datasets different from the test datasets.
 We compare methods targeting specific forgery types and generic forgery detection as follows:
DoaGan~\cite{islam2020doa}, Buster-Net~\cite{wu2018busternet} and CMSD-STRD~\cite{chen2020serial} are designed for copy-move detection.
MFCN~\cite{salloum2018image}, RRU-Net~\cite{bi2019rru} and CAT-Net~\cite{kwon2022learning} are designed for splicing detection.
HP-FCN~\cite{li2019localization}, IID-Net~\cite{wu2021iid} and TLTF-LEFF~\cite{li2023transformer} are designed for inpainting detection.
H-LSTM~\cite{bappy2019hybrid}, SPAN~\cite{hu2020span}, MVSS-Net~\cite{dong2022mvss}, SATL-Net~\cite{zhuo2022self}, PSCC-Net~\cite{liu2022pscc}, HiFi-Net~\cite{guo2023hierarchical} and IML-Vit~\cite{ma2023imlvit} are designed for generic forgery detection.

\noindent \textbf{Detection evaluation.}
Table~\ref{tab:detection result} shows the forgery detection performance. 
We observe that many methods perform very poorly, with the AUC of their performance approaching 0.5, i.e., close to random guessing, e.g., Buster-Net, H-LSTM, SPAN, and HiFi-Net.
Thanks to the proposed Re-MTKD framework, which includes Cue-Net and the Re-DTS strategy, our method achieves SOTA performance across all tampering types of datasets. 
Especially on the multi-tampering dataset, which has more complex types of tampered data, all compared methods perform poorly, but our method is 8.7\% ahead of the second place in the AUC score.

\begin{table*}[]
        \setlength\tabcolsep{1.4mm}
	\centering
        \small
        \begin{tabular}{llccccccccc}
        \hline
        \multicolumn{2}{l}{\multirow{2}{*}{\textbf{KD Strategy}}} & \multicolumn{4}{c}{\textbf{Detection}} & \multicolumn{4}{c}{\textbf{Localization}} & \multirow{2}{*}{\textbf{Average}} \\ \cline{3-6} \cline{7-10}
        \multicolumn{2}{l}{} & Copy-move & Splicing & Inpainting & Multi & Copy-move & Splicing & Inpainting & Multi &  \\
        \hline
        \multicolumn{2}{l}{0: -} & 0.873 & 0.881 & 0.794 & \underline{0.764} & \underline{0.276} & 0.611 & 0.530 & 0.419 & 0.643  \\ \hdashline
        \multicolumn{2}{l}{1: Single-Com} & 0.884 & 0.906 & 0.767 & 0.721 & 0.261 & \textbf{0.678} & 0.510 & \underline{0.436} & 0.645  \\
        \multicolumn{2}{l}{2: Single-Spl} & 0.721  & 0.883  & 0.888  & 0.699  & 0.264 & 0.637 & 0.589 & 0.390 & 0.634  \\
        \multicolumn{2}{l}{3: Single-Inp} & 0.886 & 0.901  & 0.914  & 0.696  & 0.259 & 0.652 & 0.721 & 0.372 & 0.675  \\ \hdashline
        \multicolumn{2}{l}{4: U-Ensemble} & 0.880  & 0.890  & 0.953   & 0.741  & 0.198   & 0.635 & 0.600 & 0.392 & 0.661  \\ \hdashline
        \multicolumn{2}{l}{5: Ours ($reward_{1}+\mathcal{L}_{\textbf{soft3}}$)} & 0.888  & 0.907  & 0.943  & 0.738  & 0.222 & 0.632 & 0.750 & 0.412 & 0.686  \\ 
        \multicolumn{2}{l}{6: Ours ($reward_{2}+\mathcal{L}_{\textbf{soft3}}$)} & \underline{0.890} & \textbf{0.923} & 0.949 & 0.752  & 0.235 & 0.645 & \underline{0.751} & 0.426 & \underline{0.696} \\ \hdashline
        \multicolumn{2}{l}{7: Ours ($reward_{3}+\mathcal{L}_{\textbf{soft1}}$)} & \underline{0.890} & 0.891 & 0.919 & 0.716 & 0.257 & 0.634 & 0.729 & 0.432 & 0.684 \\
        \multicolumn{2}{l}{8: Ours ($reward_{3}+\mathcal{L}_{\textbf{soft2}}$)} & 0.865 & 0.830 & \underline{0.957} & 0.694 & 0.210 & 0.639 & 0.683 & 0.398 & 0.660 \\ 
        \multicolumn{2}{l}{9: Ours ($reward_{3}+\mathcal{L}_{\textbf{soft3}}$)} & \textbf{0.900} & \underline{0.918} & \textbf{0.981} & \textbf{0.778} & \textbf{0.277} & \underline{0.676} & \textbf{0.789} & \textbf{0.444} & \textbf{0.720} \\ \hline
        \end{tabular}
        \caption{F1 performance of various KD strategies on all test sets. (1) \textbf{``-'':} Baseline, only the student model Cue-Net is directly trained on task label data without any soft target labels from teacher models. (2) \textbf{Single-Com\textbackslash Spl\textbackslash Inp:} Single forgery-specific detection teacher model for KD. (3) \textbf{U-Ensemble}~\cite{yu2023learning}: Equal weight assignment to all teacher models, with the student model receiving their average output. (4) \textbf{Ours ($reward$):} Our Re-DTS strategy, we compare three different reward functions, as detailed in Eq.~\ref{Eq:reward}. (5) \textbf{Ours ($\mathcal{L}_{\textbf{soft}}$):} Our Re-DTS strategy, we compare the different ``soft'' loss transfer by teacher models in KD, $\mathcal{L}_{\textbf{soft1}}=\mathcal{L}_{seg}$, $\mathcal{L}_{\textbf{soft2}}=\mathcal{L}_{cls}$, $\mathcal{L}_{\textbf{soft3}}=\mathcal{L}_{seg}+\mathcal{L}_{cls}$.}
        \label{tab:kd}
\end{table*}

\noindent \textbf{Localization evaluation.} 
Table~\ref{tab:localization result} shows the forgery localization performance. 
Notably, some of the forgery-specific detection methods show competitiveness on the corresponding data, e.g., Buster-Net achieves the optimal AUC scores on copy-move data, CAT-Net achieves the second-best performance on splicing data, and IID-Net achieves competitive AUC scores on inpainting data. 
Nevertheless, these methods often struggle when applied to other types of forgeries or multi-tampering data, resulting in noticeable performance drops. 
Interestingly, few generic forgery detection methods consistently demonstrate strong efficacy across both specific and multi-tampering types, underscoring the challenge of learning multiple tampering traces simultaneously.
In contrast, our method delivers superior results across all datasets and significantly outperforms the others on average, highlighting its ability to effectively capture both the commonalities and specifics of multiple tamper races.

\subsection{Ablation Study}
This subsection primarily analyzes the effectiveness of key components of the Re-MTKD framework. Table~\ref{tab:kd} presents the ablation results for the Re-DTS strategy, along with additional strategies for assigning teacher weights. Further details on ablation experiments are provided in the Appendix.

\noindent \textbf{Effectiveness of Teacher Models in KD.}
Comparing Setups \#0 - \#3,  it can be observed that the simple Cue-Net student model struggles to cope with a wide range of tampering attacks. 
In contrast, a forgery-specific detection teacher can improve the performance of the student model on the corresponding data, e.g., \textbf{Singe-Inp} achieves 14\% and 19\% improvement in forgery detection and forgery localization $F1$ performance on inpainting data compared to Setups \#0, respectively.
Comparing Setups \#0 and \#4, \textbf{U-Ensemble} provides some performance improvement on the IFDL task, but suffers from performance drops in some manipulation types, e.g., copy-move and multi-tampering.
This indicates that averaging weights across multiple teachers fails to adequately capture the distinct commonalities of different tampering operations and may not represent the most effective strategy.

\noindent \textbf{Effectiveness of Re-DTS Strategy.}
As shown by the last five setups, we compare the effects of different $rewards$ and various teacher knowledge transfer strategies $\mathcal{L}_{\textbf{soft}}$ in the Re-DTS proposed in this paper.
\textbf{(i) $rewards:$}
Comparing Setups \#5, \#6, and \#9, the performance of the student model can be better improved by adding a reward as the supervision of the knowledge transfer from the teacher model to the student model.
For Setup \#9, the final reward ($reward_{3}$) adopted in this paper, the total rewards for the ``soft'' loss of the teacher model $\mathcal{L}_{\textbf{soft}}$ and the ``hard'' loss $\mathcal{L}_{\textbf{hard}}$, as well as the performance metrics of the student model on forgery localization $F1$ and forgery detection $Acc$, effectively contribute to the overall performance of the student model. 
\textbf{(ii) $\mathcal{L}_{\textbf{soft}}:$}
Comparing Setups \#7, \#8 and \#9, 
it can be seen that the combination of specialized knowledge on forgery detection and localization transferred by the teacher model improves the IFDL performance of the student model more efficiently.
Our method (Setup \#9) ultimately achieves excellent results not only on multiple specific forgery data, but also on the more challenging multi-tampering data, where equally efficient performance is achieved.

\begin{figure}
	\centering
	\subfloat[Only Cue-Net]{
		\includegraphics[width=0.299\linewidth]{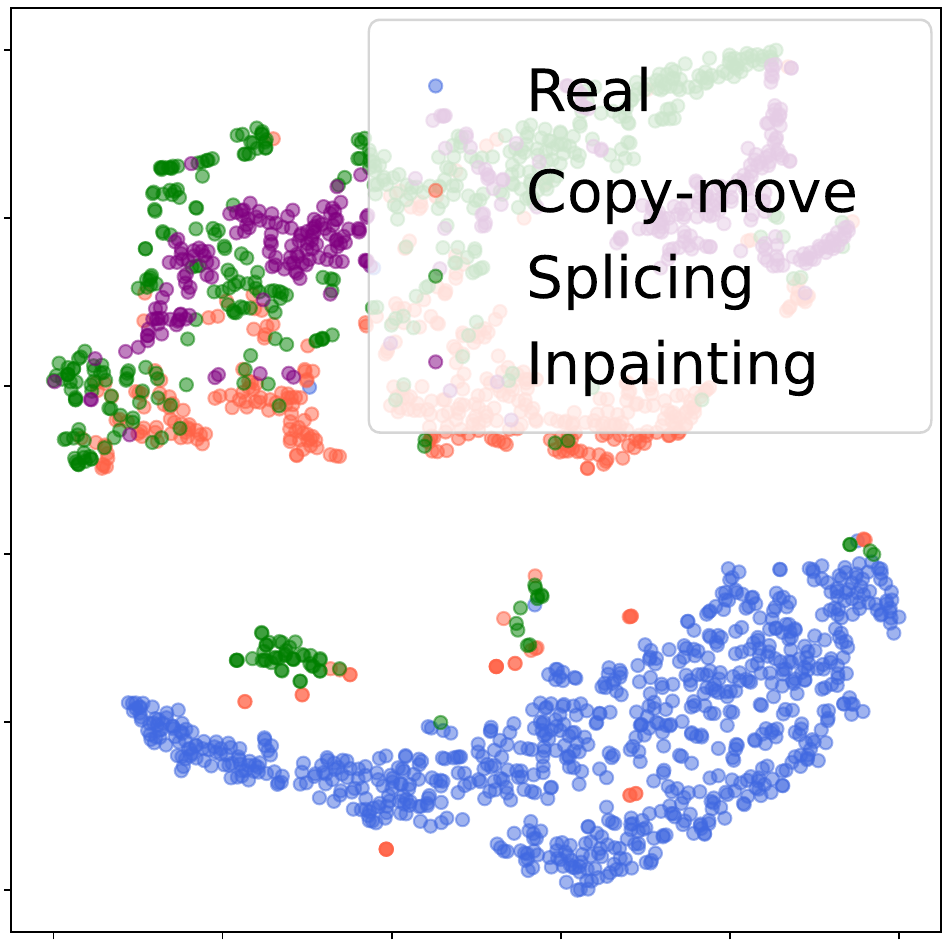}
	}
\hspace{0.01\linewidth}
 	\subfloat[U-Ensemble]{
		\includegraphics[width=0.299\linewidth]{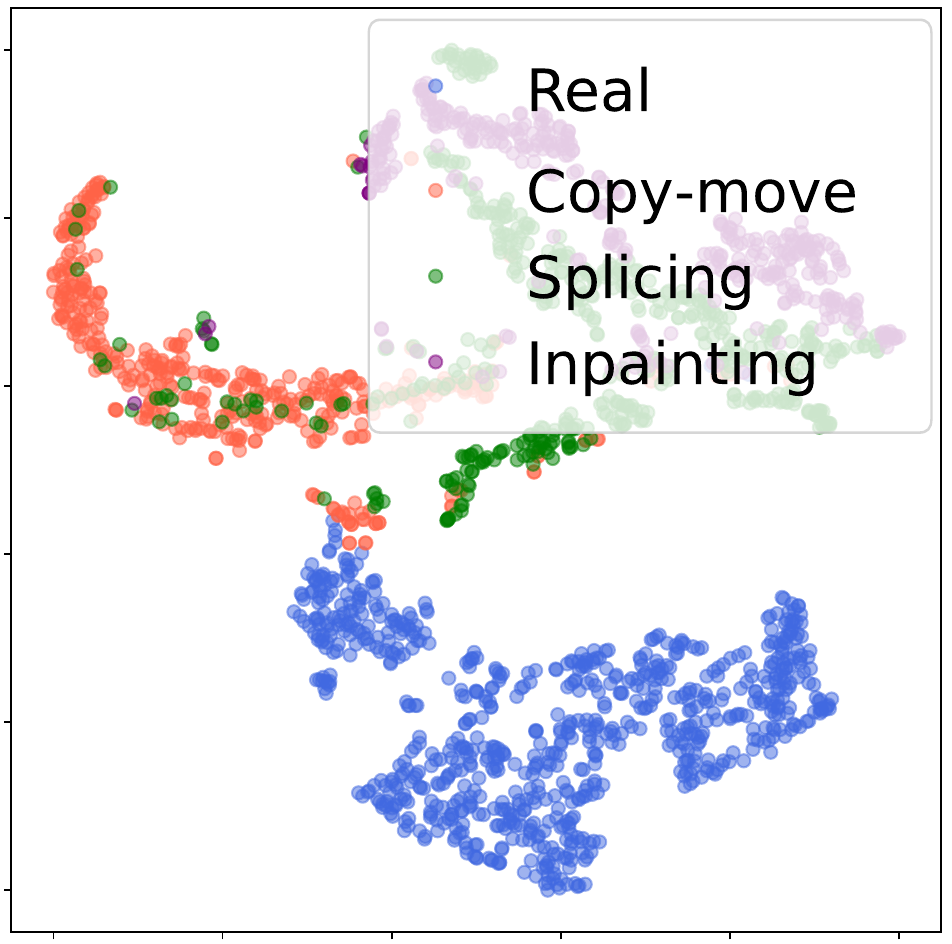}
	}
\hspace{0.01\linewidth}
 	\subfloat[Re-MTKD (Ours)]{
		\includegraphics[width=0.299\linewidth]{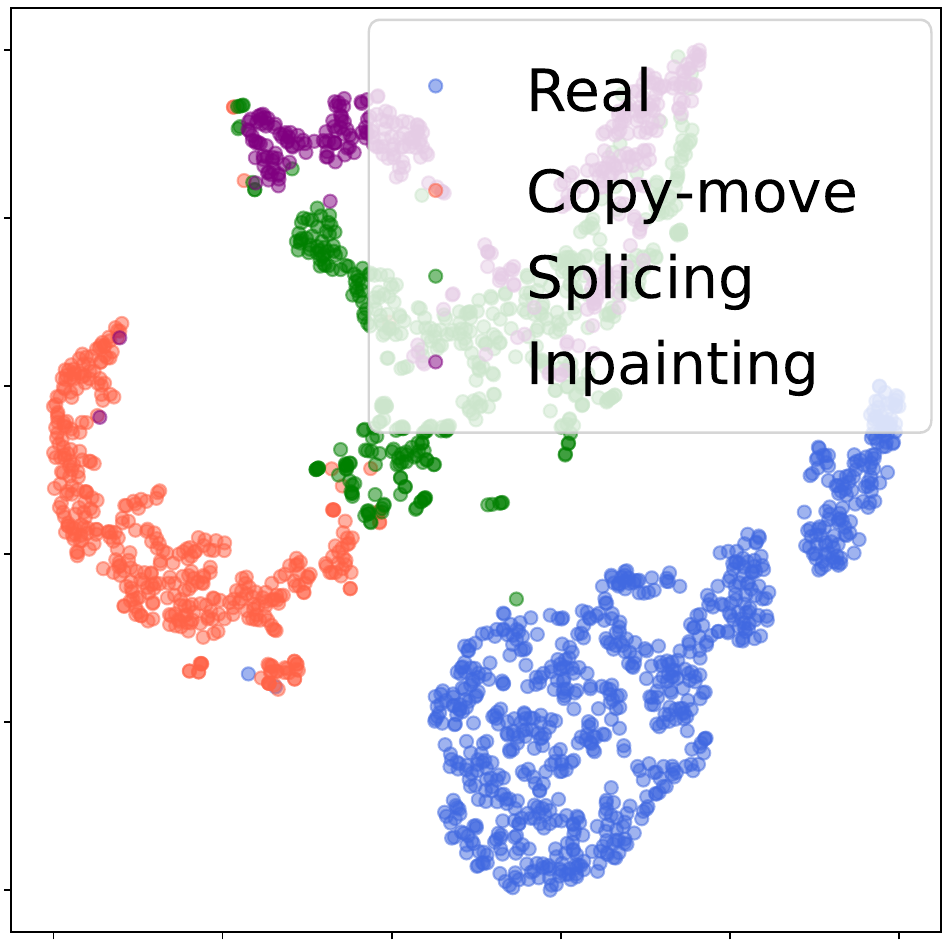}
	}
	\caption{Feature space visualization of different KD strategies on CASIA v1+ and DiverseInp.}
         \label{fig:tsne}
\end{figure}

Furthermore, we show the embedding space of learned features with t-SNE~\cite{van2008visualizing} visualization of different KD strategies in Fig.~\ref{fig:tsne}.
We observe that in our proposed Re-MTKD framework, Cue-Net, when trained with the Re-DTS strategy, effectively discriminates between the feature distributions of real and tampered samples, outperforming other KD strategies.
This demonstrates the model's ability to learn common features across various tampering types, enabling it to accurately classify all tampered samples as tampered, irrespective of the specific tampering technique. Additionally, the model also captures the specific features of each tampering type, resulting in more clustered and distinct distributions for each category of tampered samples.

\section{Conclusion}
In this paper, we propose a novel Reinforced Multi-teacher Knowledge Distillation (Re-MTKD) framework designed for image forgery detection and localization (IFDL). 
Specifically, we develop a new network called Cue-Net, featuring a ConvNeXt-UPerNet structure equipped with an Edge-Aware Module, which serves as an effective backbone for the IFDL task. 
We further introduce a Reinforced Dynamic Teacher Selection (Re-DTS) strategy, which dynamically selects specialized teacher models based on different types of tampering data, guiding the student model to effectively learn both the commonalities and specifics of various tampering traces. 
Extensive experiments across several IFDL tasks demonstrate the superior performance of our proposed method compared to existing state-of-the-art approaches.

\section{Acknowledgments}
This work was partially supported by the National Natural Science Foundation of China (Grants No.~U23B2022, U22A2030), and the Guangdong Major Project of Basic and Applied Basic Research (Grand No.~2023B0303000010).

\bibliography{aaai25}
\clearpage
\section{Supplementary Materials}
In this supplementary material, we provide more details of our work, including: 
1) More Experimental Setup: details of the dataset settings and evaluation metrics.
2) More Experimental Results: additional quantitative and qualitative experiments, robustness test results, as well as inference efficiency, comparing our method with SOTA methods.
3) More Ablation Studies: further ablation experiments on the Cue-Net backbone and the Reinforced Dynamic Teacher Selection (Re-DTS) strategy.
\section{More Experimental Setup}
\noindent \textbf{Datasets Details.}
Details of the datasets used for training and testing are summarized in Table~\ref{tab:dataset}.
\textbf{For Copy-move (Com)},
we use CASIA v2~\cite{dong2013casia} and Tampered Coco~\cite{liu2022pscc} for training and CASIA v1+~\cite{dong2013casia}, NIST16~\cite{nimble2016datasets} and Coverage~\cite{wen2016coverage} for testing.
\textbf{For Splicing (Spl)}, we use CASIA v2~\cite{dong2013casia} and Fantastic-Reality~\cite{kniaz2019point} for training, and CASIA v1+~\cite{dong2013casia}, NIST16~\cite{nimble2016datasets}, Columbia~\cite{hsu2006detecting} and DSO-1~\cite{de2013exposing} for testing.
\textbf{For Inpainting (Inp)},
we use GC Dresden \& Places~\cite{wu2021iid} for training, and NIST16~\cite{nimble2016datasets}, AutoSplicing~\cite{jia2023autosplice} and DiverseInp~\cite{wu2021iid} for testing.
\textbf{For Multi-tampering (Multi)},
as this type involves composite operations of single tampering types, 
we only use IFC~\cite{IFC}, Korus~\cite{korus2016multi}, and IMD2020~\cite{novozamsky2020imd2020} for testing.
Examples of the four types of tampering operations described previously are illustrated in Fig.~\ref{tab:dataset}.

\begin{itemize}
     \item \textbf{CASIA v2 \& v1+ }~\cite{dong2013casia} contains copy-move and splicing images in which tampered regions are carefully selected and complemented by various post-processing techniques, including filtering and blurring. CASIA~\cite{dong2013casia} is divided into CASIA v2 (7,491 real samples and 5,123 tampered samples) for training and CASIA v1 (800 real samples and 920 tampered samples) for testing.
     To prevent data overlap between real and tampered images, we adopt the approach outlined in~\cite{dong2022mvss}, incorporating COREL~\cite{wang2001simplicity} data as real images into CASIA v1+.
     \item \textbf{Fantasitic Reality}~\cite{kniaz2019point} is a spliced image dataset that is more extensive in terms of scene diversity and image count.
     For training, we sample 3,630 real images and 19,432 tampered images from this dataset.
    \begin{table}[h]
            \small
            \setlength\tabcolsep{1mm}
            \setlength{\aboverulesep}{0.6pt}
            \setlength{\belowrulesep}{0.6pt}
            \begin{tabular}{lcccccc}
    			\toprule
    			\textbf{Dataset} & \textbf{Neg.} & \textbf{Pos.} & \textbf{Com.} & \textbf{Spl.} & \textbf{Inp.} & \textbf{Multi.}       \\ \midrule
    			\multicolumn{7}{l}{\textbf{\# Training}}                         \\ \midrule
    			CAISA v2          & 7,491 & 5,123  & 3,295  & 1,828  & -     & -    \\
    			Fantastic Reality & 3,630 & 19,423 & -     & 19,423 & -     & -    \\
    			Tampered Coco     & -    & 10,000 & 10,000 & -     & -     & -    \\ 
    			GC Dresden       & -    & 7,500  & -     & -     & 7,500  & -    \\ 
    			GC Places         & -    & 7,500  & -     & -     & 7,500  & -    \\ \midrule
    			\multicolumn{7}{l}{\textbf{\# Testing}}                          \\ \midrule
    			CAISA v1+         & 800  & 920   & 459   & 461   & -     & -    \\
    			NIST16            & -    & 564   & 68    & 288   & 208   & -    \\
    			Coverage          & 100  & 100   & 100   & -     & -     & -    \\
    			Columbia          & 183  & 180   & -     & 180   & -     & -    \\
    			DSO-1             & -    & 100   & -     & 100   & -     & -    \\
    			AutoSplicing      & 2,273 & 3,621  & -     & -    & 3,621    & -    \\
    			DiverseInp        & 2,000 & 2,200  & -     & -     & 2,200  & -    \\
    			IFC              & 1,050 & 442   & -     & -     & -     & 442  \\
    			Korus             & 220  & 220   & -     & -     & -     & 220  \\
    			IMD2020           & 414  & 2,010  & -     & -     & -     & 2,010 \\ \bottomrule
    		\end{tabular}
     	\caption{Summary of IFDL datasets involved in this paper. 
       Neg denotes the negative sample, i.e., the real image. Pos denotes the positive sample, i.e., the tampered image.
     Com, Spl, and Inp indicate three common image manipulation types: copy-move, splicing, and inpainting. Multi represents a multi-tampering operation, which is a combination of the above three tampering operations.}
     \label{tab:dataset}
    \end{table}
     \item \textbf{Tampered Coco}~\cite{liu2022pscc} is a synthetic dataset leveraging MS COCO~\cite{lin2014microsoft} as its image source. 
     It contains three types of tampering, i.e., copy-move, splicing, and inpainting.
     For training, we employ 10,000 copy-move images from this dataset.
     \item \textbf{GC Dresden \& Places}~\cite{wu2021iid} uses the Places~\cite{zhou2017places} (JPEG lossy compression) and Dresden~\cite{gloe2010dresden} (NEF lossless compression) datasets as the base images,
     employing the inpainting method~\cite{liu2018image} to generate 48,000 images. For training, we employ 7,500 images in each dataset.
     \item \textbf{NIST16}~\cite{nimble2016datasets} presents a challenging collection encompassing copy-move, splicing and inpainting techniques. The manipulations included in this selection are post-processed to mask visible traces.
     \item \textbf{Coverage}~\cite{wen2016coverage} contains 100 tampered images manipulated by copy-moving. Each image undergoes post-processing to remove visual traces.
      \item \textbf{Columbia}~\cite{hsu2006detecting} contains 180 uncompressed spliced tampered images and 183 real images.
     \item \textbf{DSO-1}~\cite{de2013exposing} contains 100 images of people undergoing splicing tampering operations.
     \item \textbf{AutoSplicing}~\cite{jia2023autosplice} uses the language-image model based on the diffusion model, DALL-E2~\cite{ramesh2022hierarchical},  to locally or globally modify images guided by text prompts. 
     The dataset comprises 3,621 tampered images and 2,273 real images, with varying dimensions from 256 × 256 to 4,232 × 4,232 pixels.
     Despite the inclusion of ``splicing'' in its name, we categorize it as an inpainting dataset.
     \item \textbf{DiverseInp}~\cite{wu2021iid} undergoes tampering using 10 inpainting methods (6 DL-based and 4 traditional) on CelebA~\cite{karras2017progressive} and ImageNet~\cite{deng2009imagenet}. Each inpainting method contributes 1,000 images, from which we sample 220 for testing.
     \item \textbf{IFC} ~\cite{IFC} originates from the first forensic challenge organized by IFS-TC and contains 1,050 real and 442 tampered images. 
     It mirrors real-life tampering scenarios without distinguishing between tampering operations.
    \item \textbf{Korus}~\cite{korus2016multi} contains 220 images capturing daily scenes from four digital cameras. Some tampered images undergo editing using multiple tampering techniques.
     \item \textbf{IMD2020}~\cite{novozamsky2020imd2020} comprises 2,010 real-life manipulated images sourced from the Internet along with 414 corresponding real images.
\end{itemize}

\noindent \textbf{Evaluation metrics.}
In the main paper, we evaluate our model using several key metrics. 
For forgery localization, we report pixel-level F1, Intersection over Union (IoU) and Area Under the Curve (AUC). For forgery detection, in addition to image-level Accuracy (Acc), F1 and AUC. For both forgery detection and localization, the default threshold is 0.5 unless otherwise specified.
\begin{figure}[tb] 
        \centering
	\includegraphics[width=0.47\textwidth]{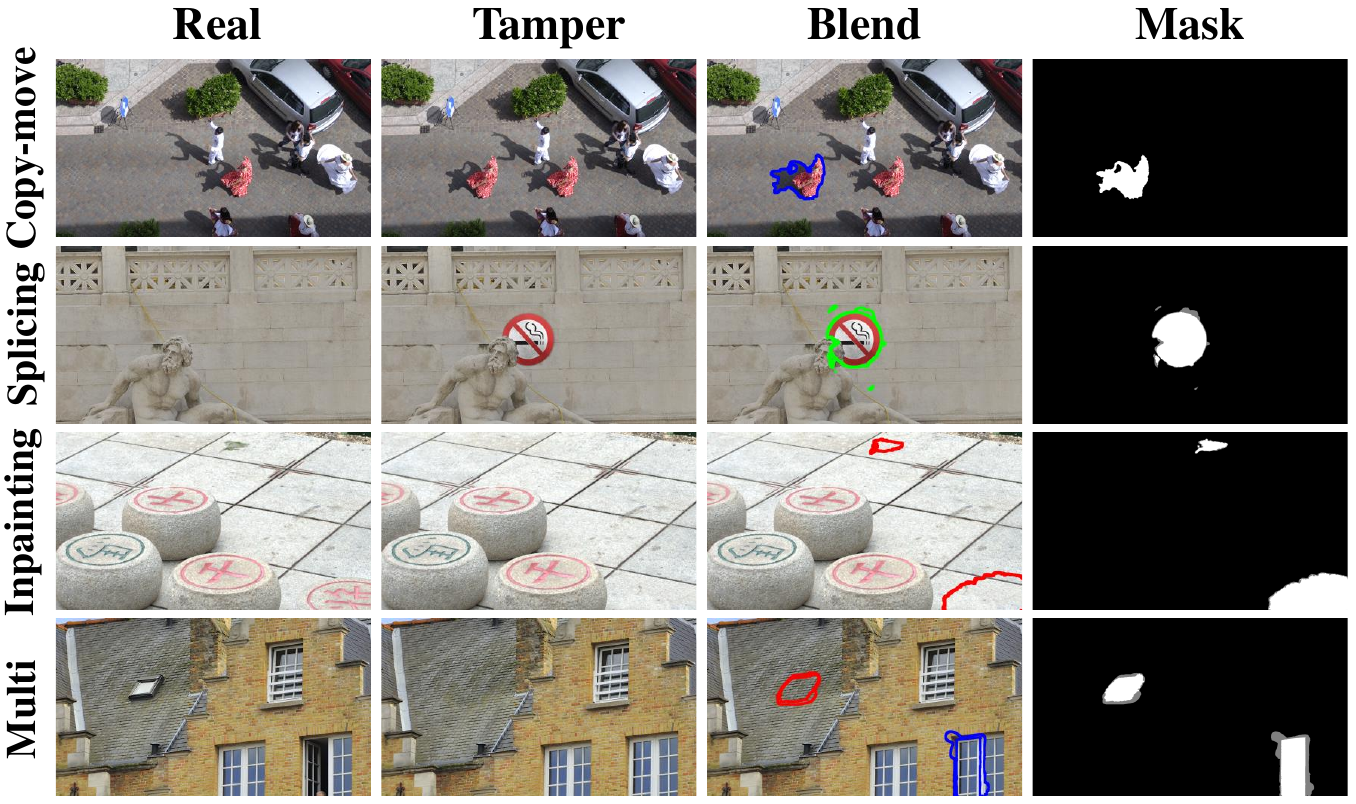}
	\caption{Four examples are copy-move (copying and moving an object within the target image), splicing (pasting the object from the source image to the target image), inpainting (erasing the object from the target image) and multi-tampering (possible combinations of the above three tampering operations), respectively. In column 3, we emphasize the tampered regions of copy-move, splicing and inpainting using \textbf{blue}, \textbf{green} and \textbf{red} edges, respectively. In this case, the multi-tampered image contains both copy-move and inpainting tampering combinations, which presents a significant challenge to the IFDL task.} \label{fig:pre-task}
\end{figure}

\begin{table*}[]
\centering
\small
\setlength{\aboverulesep}{0.6pt}
\setlength{\belowrulesep}{0.6pt}
\setlength\tabcolsep{1.6mm}
\resizebox{\textwidth}{!}{
\begin{tabular}{clcccccccccc}
\toprule
\multicolumn{2}{c}{\multirow{2}{*}{\textbf{Methods}}} & \multirow{2}{*}{\textbf{\#Data}} & \multicolumn{4}{c}{\textbf{Detection}} & \multicolumn{4}{c}{\textbf{Localization}} & \multirow{2}{*}{\textbf{Average}} \\ \cmidrule(lr){4-7} \cmidrule(lr){8-11}
\multicolumn{2}{c}{} &  & Copy-move & Splicing & Inpainting & Multi & Copy-move & Splicing & Inpainting & Multi &  \\ \midrule
\multirow{3}{*}{\rotatebox{90}{\centering D-Com}} & DoaGan\footnotemark[1] & 160k & 0.756 & 0.516 & 0.522 & 0.560 & 0.785 & 0.550 & 0.561 & 0.569 & 0.602 \\
 & Buster-Net\footnotemark[1] & 81.7k & 0.500 & 0.500 & 0.500 & 0.500 & 0.806 & 0.715 & 0.666 & 0.705 & 0.612 \\
 & CMSD-STRD\footnotemark[2] & - & - & - & - & - & - & - & - & - & - \\ \midrule
\multirow{3}{*}{\rotatebox{90}{\centering D-Spl}} & RRU-Net & 5.3k & 0.474 & 0.598 & 0.519 & 0.501 & 0.577 & 0.922 & 0.531 & 0.692 & 0.602 \\
 & CAT-Net v1 & 960k & 0.492 & 0.830 & 0.500 & 0.782 & 0.586 & 0.833 & 0.636 & 0.826 & 0.686 \\
 & CAT-Net v2 & 876k & 0.487 & 0.729 & 0.608 & 0.616 & 0.594 & 0.770 & 0.630 & 0.721 & 0.644 \\ \midrule
\multirow{3}{*}{\rotatebox{90}{\centering D-Inp}} & HP-FCN & 50k & 0.735 & 0.442 & \underline{0.928} & 0.621 & 0.494 & 0.605 & 0.557 & 0.392 & 0.597 \\
 & IID-Net & 48k & 0.648 & 0.795 & 0.858 & 0.703 & 0.809 & 0.888 & 0.748 & 0.639 & 0.761 \\
 & TLTF-LEFF & 48k & 0.541 & 0.445 & 0.587 & 0.497 & 0.484 & 0.489 & 0.641 & 0.489 & 0.522 \\ \midrule
\multirow{8}{*}{\rotatebox{90}{\centering D-Generic}} & ManTra-Net & 102k & 0.534 & 0.589 & 0.474 & 0.665 & 0.624 & 0.733 & 0.645 & 0.753 & 0.627 \\
 & SPAN & 102k & 0.500 & 0.500 & 0.500 & 0.500 & 0.650 & 0.722 & 0.501 & 0.681 & 0.569 \\
 & MVSS-Net & 13k & \underline{0.895} & \textbf{0.973} & 0.852 & 0.685 & \underline{0.830} & 0.890 & 0.813 & 0.773 & 0.839 \\
 & SATL-Net & 88.9k & 0.377 & 0.623 & 0.533 & 0.556 & 0.577 & 0.671 & 0.573 & 0.544 & 0.557 \\
 & PSCC-Net & 100k & 0.440 & 0.619 & 0.498 & 0.402 & 0.625 & 0.747 & 0.580 & 0.767 & 0.585 \\
 & HiFi-Net & 1,725k & 0.256 & 0.133 & 0.755 & 0.336 & 0.464 & 0.483 & 0.517 & 0.481 & 0.428 \\
 & TruFor\footnotemark[3] & \multirow{2}{*}{876k} & \multirow{2}{*}{0.759} & \multirow{2}{*}{0.948} & \multirow{2}{*}{0.740} & \textbf{0.819} & \multirow{2}{*}{\textbf{0.895}} &  \multirow{2}{*}{\underline{0.938}} & \multirow{2}{*}{0.847} & \textbf{0.886} & 0.854 \\ 
 & TruFor\textsuperscript{$\dag$} &  &  &  &  & 0.595 &  &   &  & 0.686 & 0.801 \\
 & IML-Vit & 13k & 0.692 & 0.766 & 0.577 & 0.661 & 0.849 &  0.797 & 0.738 & 0.769 & 0.731 \\ 
 & DiffForensics & 32k & 0.863 & 0.955 & \underline{0.947} & 0.745 & 0.809 & \textbf{0.939} & \underline{0.907} & 0.855 & \underline{0.878} \\ \midrule
 & Ours & 60k & \textbf{0.952} & \underline{0.970} & \textbf{0.994} & \underline{0.790} & 0.764 & 0.934 & \textbf{0.965} & \underline{0.858} & \textbf{0.903} \\ \bottomrule
\end{tabular}}
\caption{Image-level and Pixel-level AUC performance of image forgery detection and localization. The \textbf{best} and \underline{2nd-best} results are highlighted. All methods utilize the pre-trained models from the original papers.}
\label{tab: supplement_detection_localization}
\end{table*}

\section{More Experimental Results}
\subsection{Comparison with the State-of-the-Art Methods}
In this section, we present additional quantitative, qualitative and robustnes experiments, as well as inference efficiency.

\noindent \textbf{Quantitative Results.} 
We provide a more comprehensive overview of the performance of the SOTA methods utilized in the main paper. 
The IFDL results obtained using the \textbf{official pre-trained models} of these SOTA methods are presented in Table~\ref{tab: supplement_detection_localization}.
It can be observed that specific forgery detection methods exhibit better performance on the corresponding tampering type data. For example, DoaGan~\cite{islam2020doa} and Buster-Net~\cite{wu2018busternet} show exceptional localization performance on copy-move data, while RRU-Net~\cite{bi2019rru} and CAT-Net v1~\cite{kwon2021cat} show superior localization capabilities on splicing data. Moreover, HP-FCN~\cite{li2019localization} and TLTF-LEFF~\cite{li2023transformer} show remarkable detection and localization performance respectively on inpainting data.
However, many generic forgery detection methods, such as SPAN~\cite{hu2020span}, SATL-Net~\cite{zhuo2022self}, PSCC-Net~\cite{liu2022pscc}, and Hi-Fi-Net~\cite{guo2023hierarchical}, perform poorly across multiple tamper types, despite being trained on large datasets.

Additionally, we include competitive comparison methods, ManTra-Net~\cite{wu2019mantra} and TruFor~\cite{guillaro2023trufor}, but they are not covered in the main paper. This is because neither method has publicly released training code, and only test code is available.
Notably, TruFor is trained on larger-scale data (TruFor with dataset of 867k $vs$ our method with dataset of 60k) and \textbf{has been affected by data leakage} (TruFor is trained using IMD2020~\cite{novozamsky2020imd2020}, with this paper serving as part of the multi-tampering type test data.).
In particular, TruFor demonstrates exceptional forgery localization performance on copy-move and splicing data. 
However, its forgery detection performance is suboptimal, and it also shows a notable degradation on inpainting data, highlighting the challenge of handling multiple tampered data types simultaneously.
For multi-tampering data, our method still achieves comparable performance to TruFor,
even though TruFor's test model suffers from data leakage as described above.
When the IMD2020 dataset, which has data leakage issues, is excluded, our method outperforms TruFor by 10.2\% in average AUC score.
Overall, the Re-MTKD framework proposed in this paper effectively improves the model's performance across various tampered data types and achieves optimal average AUC performance.

\footnotetext[1]{DoaGan~\cite{islam2020doa} and Buster-Net~\cite{wu2018busternet} only have test code available. Generally, these two copy-move detection methods require first locating the tampered region and then distinguishing between the source and target parts, making them suitable only for copy-move data. In this paper, to provide a comprehensive overview for readers, we also report the test results on other types of tampered data.}
\footnotetext[2]{
The original CMSD-STRD~\cite{chen2020serial} is a three-class classification task, making it unsuitable for calculating binary classification AUC performance for IFDL task. In the main paper, we adapt the backbone network of this method for the IFDL task.}
\footnotetext[3]{$\dag$ indicates the non-data-leakage version of the test results, where the IMD2020~\cite{novozamsky2020imd2020} has been excluded from the multi-tampering data.}

\begin{figure}[tb] 
        \centering
	\includegraphics[width=0.47\textwidth] {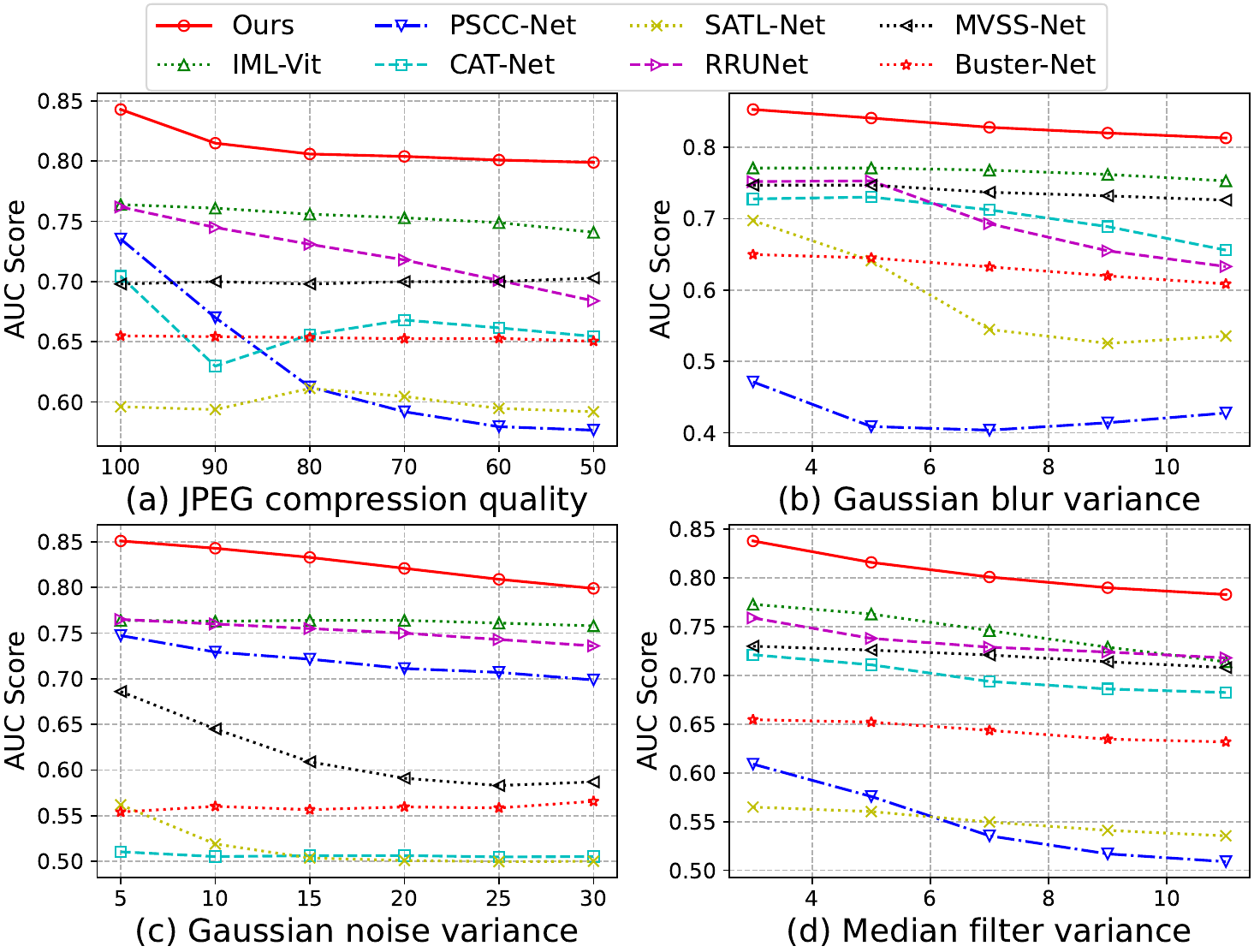}
	\caption{Robustness against JPEG compression, Gaussian blur, Gaussian noise and Median filtering effects.} 
        \label{fig:robust_result}
\end{figure}

\begin{figure*}[tb] 
        \centering
	\includegraphics[width=0.99\textwidth] {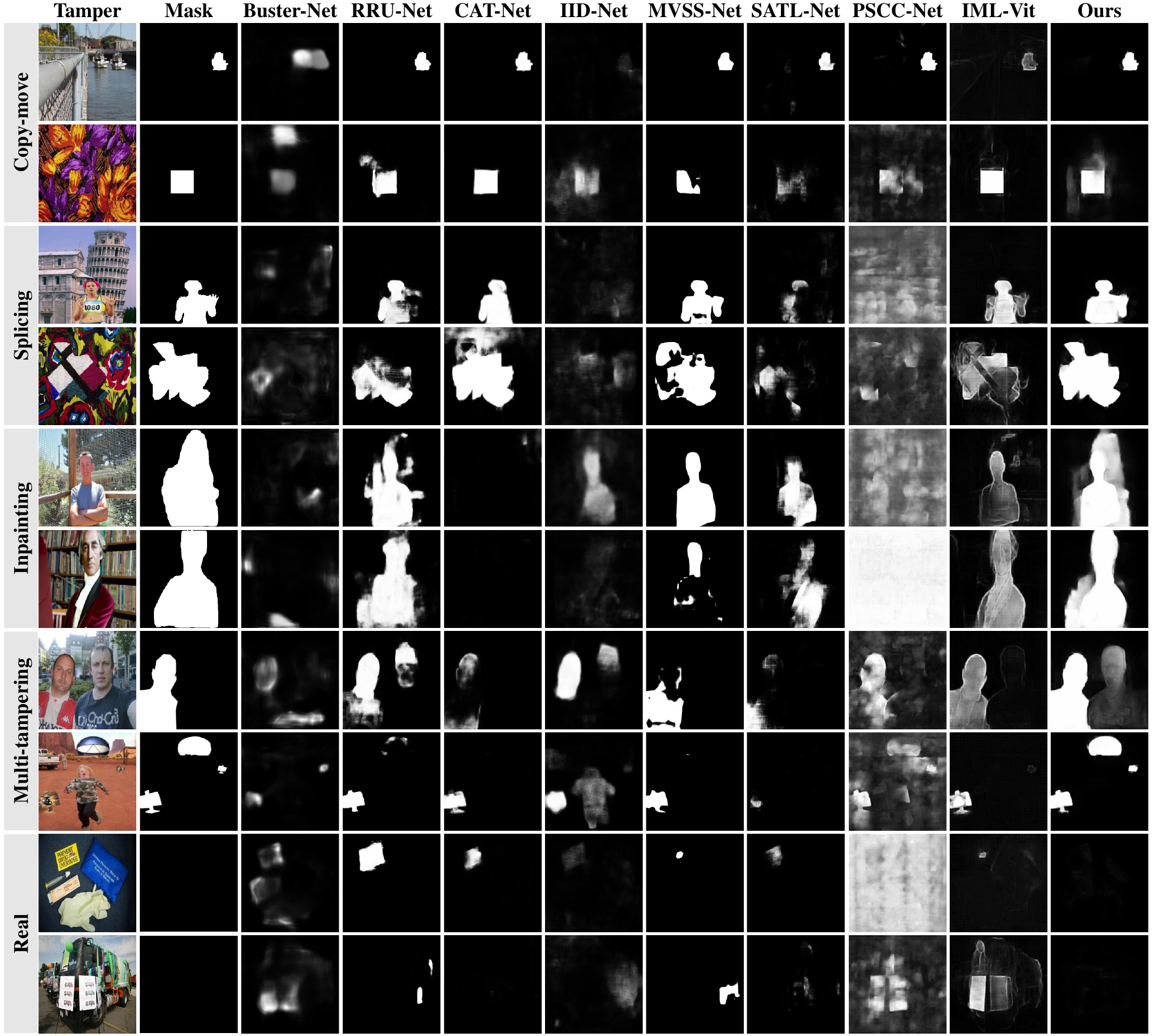}
 	\caption{Qualitative localization evaluations on tampered data, including copy-move, splicing, inpainting, and multi-tampering data, as well as real data.} 
        \label{fig:main_test}
\end{figure*}

\noindent \textbf{Qualitative Results.}
We present  the qualitative evaluations of both specific and generic forgery detection methods on various types of tampered data, including copy-move, splicing, inpainting, and multi-tampering data, as well as on real data, as shown in Fig.~\ref{fig:main_test}. 
It can be seen that specific forgery detection methods perform better on the corresponding data, such as Buster-Net~\cite{wu2018busternet} on copy-move data, RRU-Net~\cite{bi2019rru} and CAT-Net~\cite{kwon2022learning} on splicing data, and IID-Net~\cite{wu2021iid} on inpainting data. 
Compared to generic forgery detection methods, our approach ultimately achieves more accurate forgery localization results across various types of tampered data. Additionally, our proposed method demonstrates a lower false positive rate when applied to real images.

\noindent \textbf{Robustness.} 
We further evaluated the robustness of localization results when facing common image perturbations in social media laundering, i.e., JPEG compression, Gaussian blurring, Gaussian noising and Median filtering.
As shown in Fig.~\ref{fig:robust_result},
some generic IFDL methods, such as PSCC-Net~\cite{liu2022pscc} against JPEG compression and SATL-Net~\cite{zhuo2022self} against Gaussian blurring, suffer significant performance degradation.
In contrast, our method demonstrates substantial robustness in localization performance across the full range of post-processing attacks.

\begin{table}[]
\centering
\small
\setlength{\aboverulesep}{0.6pt}
\setlength{\belowrulesep}{0.6pt}
\setlength\tabcolsep{0.8mm}
\resizebox{0.48\textwidth}{!}{ 
\begin{tabular}{lccc}
\toprule
\textbf{Methods} & \textbf{Speed (ms/img) $\downarrow$} & \textbf{Memory (MB) $\downarrow$} & \textbf{AUC (\%) $\uparrow$} \\ \midrule
MVSS-Net & 129.37 & 4293 & 0.839 \\
PSCC-Net & 64.49 & 5159 & 0.585 \\
TruFor & 64.87 & \underline{3115} & 0.854 \\
DiffForensics & \underline{59.5} & \textbf{2377} & \underline{0.878} \\
Ours & \textbf{38.2} & 3147 & \textbf{0.903} \\ \bottomrule
\end{tabular}}
\caption{Comparison of inference memory usage, speed, and average AUC across different SOTA methods. The AUC values are from Table~\ref{tab: supplement_detection_localization}, representing the average performance on image forgery detection and localization.}
\end{table}

\noindent \textbf{Inference Efficiency.} 
As shown in Table 6, we selected competitive SOTA methods for testing on the copy-move dataset, where our method achieved the fastest inference speed and competitive memory usage, averaged over five runs. Our method demonstrated superior inference efficiency, with an inference speed of 38.2 ms/img, significantly outperforming other methods such as MVSS-Net (129.37 ms/img) and TruFor (64.87 ms/img). Ultimately, it achieved the highest AUC of 0.903, confirming that our method strikes the optimal balance between inference efficiency and performance, excelling in both detection and localization tasks.

\section{More Ablation Studies}
In this section, we conduct additional ablation studies of several components for the Cue-Net backbone and Reinforced Dynamic Teacher Selection strategy (Re-DTS).

\begin{table}[tb]
\small
\centering
\setlength{\aboverulesep}{0.6pt}
\setlength{\belowrulesep}{0.6pt}
\setlength\tabcolsep{1.1mm}
\resizebox{0.48\textwidth}{!}{ 
\begin{tabular}{c@{\hspace{0.9\tabcolsep}}c@{\hspace{0.9\tabcolsep}}c@{\hspace{0.9\tabcolsep}}c@{\hspace{0.9\tabcolsep}}cccccc}
\toprule
\multirow{2}{*}{\textbf{PPM}} & \multicolumn{3}{c}{\textbf{EAM}} & \multicolumn{2}{c}{\textbf{Detection}} & \multicolumn{2}{c}{\textbf{Localization}} & \multicolumn{2}{c}{\textbf{Average}} \\ 
\cmidrule(lr){2-4} \cmidrule(lr){5-6} \cmidrule(lr){7-8} \cmidrule(lr){9-10}
 &low  & hight & fusion & F1 & AUC & F1 & AUC & F1 & AUC \\ \midrule
- & - & - & - & .683 & .769 & .370 & .864 & .526 & .817 \\
$\checkmark$ & - & - & - & .710 & \underline{.782} & \underline{.404} & .863 & .557 & \underline{.822} \\
- & - & - & $\checkmark$ & .707 & .752 & .394 & \underline{.870} & .551 & .811 \\
$\checkmark$ & $\checkmark$ & - & - & .705 & .729 & .402 & .866 & .553 & .797 \\
$\checkmark$ & - & $\checkmark$ & - & \underline{.739} & .754 & .395 & \textbf{.876} & \underline{.567} & .815 \\
$\checkmark$ & - & - & $\checkmark$ & \textbf{.764} & \textbf{.807} & \textbf{.419} & .868 & \textbf{.591} & \textbf{.837} \\ \bottomrule
\end{tabular}}
\caption{Ablation Study of Cue-Net Backbone. Test on Multi-tampering type data. Note that EAM's contribution is reflected through $\mathcal{L}_{edg}$.}
\label{tab: EAM}
\end{table}

\begin{figure}[] 
\centering
\includegraphics[width=0.47\textwidth]{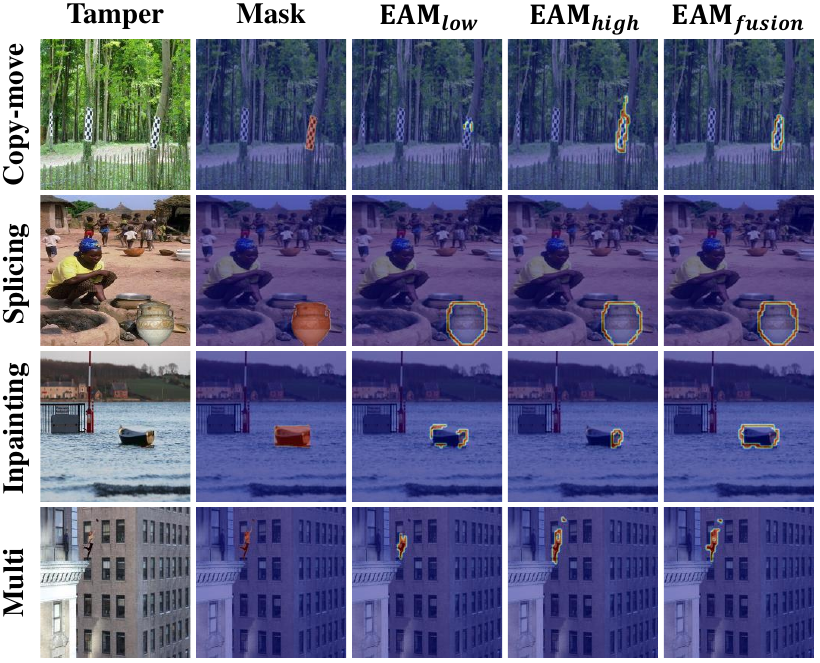}
\caption{Visualization of Edge-Aware Module results. For better visualization, we have highlighted the predictions on the image, and the pseudo-colored part of the image is the prediction result.} 
\label{fig:abtest_edge}
\end{figure}

\subsection{Cue-Net Backbone}
As shown in Table~\ref{tab: EAM},
we analyze the contribution of the low-level and high-level features of the EAM  to $\mathcal{L}_{edg}$, as well as the impact of the PPM on the overall model performance.
Comparing row 1 with row 2, and row 3 with the last row, it can be observed that the multi-scale feature aggregation of PPM improves the IFDL performance.
By comparing rows 4 and 5, it can be observed that the EAM with low-level features effectively facilitates the forgery localization performance, while the EAM with high-level features struggles to effectively determine the edges of the tampered image due to its low resolution.
Furthermore, the fusion of low-level and high-level features enables efficient learning of tampered edge traces for excellent IFDL performance.

Furthermore, we present the qualitative results of the Edge-Aware Module (EAM) in Cue-Net in Fig.~\ref{fig:abtest_edge}.
Moving from left to right, 
we observe that low-level features contain less information, making it challenging to identify tampered regions. High-level features often exhibit incomplete recognition of tampered region edges due to their lower resolution. Ultimately, the fusion of low-level and high-level features with edge supervision enhances the localization of tampered region contours.

\subsection{Reinforced Dynamic Teacher Selection strategy}
To further explore the dynamic selection of teachers in reinforced multi-teacher knowledge distillation, we present the test results for various specialized teacher models, Cue-Net, and Cue-Net trained with the Re-DTS strategy. Additionally, we visualize the dynamic selection of teacher models under different settings.

\begin{table}[tb]
\small
\centering
\setlength{\aboverulesep}{0.6pt}
\setlength{\belowrulesep}{0.6pt}
\setlength\tabcolsep{1.1mm}
\resizebox{0.48\textwidth}{!}{ 
\begin{tabular}
{lcccc}
\toprule
\textbf{Model} & \textbf{Copy-move} & \textbf{Splicing} & \textbf{Inpainting} & \textbf{Average} \\ \midrule
Teacher-Com &  (\textbf{0.394}) & 0.655 & 0.374 &  (0.430) \\
Teacher-Spl & 0.186 &  (\underline{0.674}) & 0.245 & 0.378 \\
Teacher-Inp & 0.000 & 0.018 &   (0.434) & 0.114 \\ \midrule
Only Cue-Net & 0.276 & 0.611 & \underline{0.530}  & \underline{0.459} \\ 
Re-MTKD (Ours) & \underline{0.277} & \textbf{0.676} & \textbf{0.789}  & \textbf{0.581} \\ 
\bottomrule
\end{tabular}}
\caption{Test results of different models on data of varying tampering types. Teacher-Com\textbackslash Spl\textbackslash Inp are trained on the corresponding types of tampered data. Cue-Net and Cue-Net with Re-DTS strategy (Ours) are trained on all mixed single-tampered data.
The \textbf{best} and \underline{2nd-best} results are highlighted. The $(\cdot)$ serves to highlight the optimal test results for each specific forgery teacher.}
\label{tab:single_teacher}
\end{table}

\begin{figure}[tb]
    \centering
    \subfloat[Ours: ($reward_{3}+\mathcal{L}_{\textbf{soft3}}$)]{
    \includegraphics[width=0.47\linewidth]
    {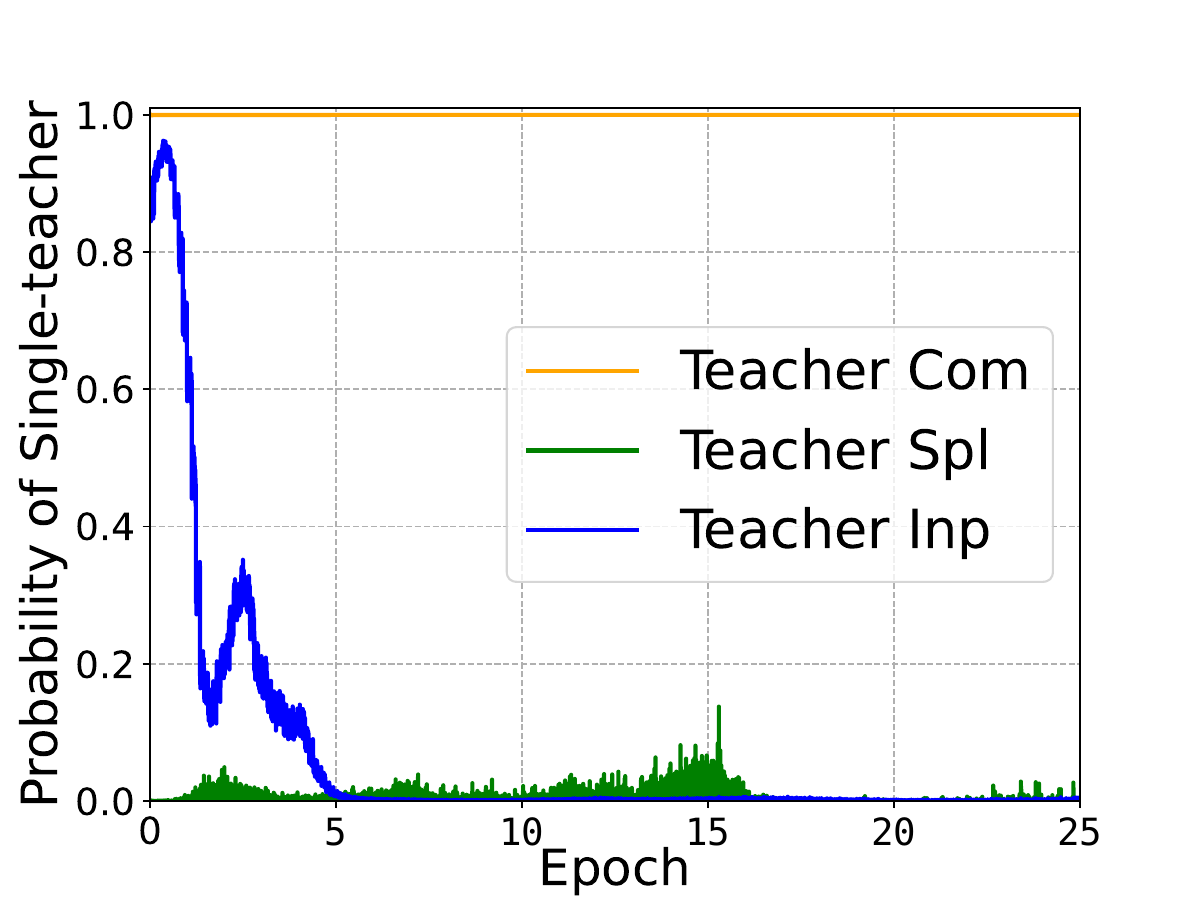}}
\hspace{0.001\linewidth}
    \subfloat[Setup \#5: ($reward_{1}+\mathcal{L}_{\textbf{soft3}}$)]{
    \includegraphics[width=0.47\linewidth]
    {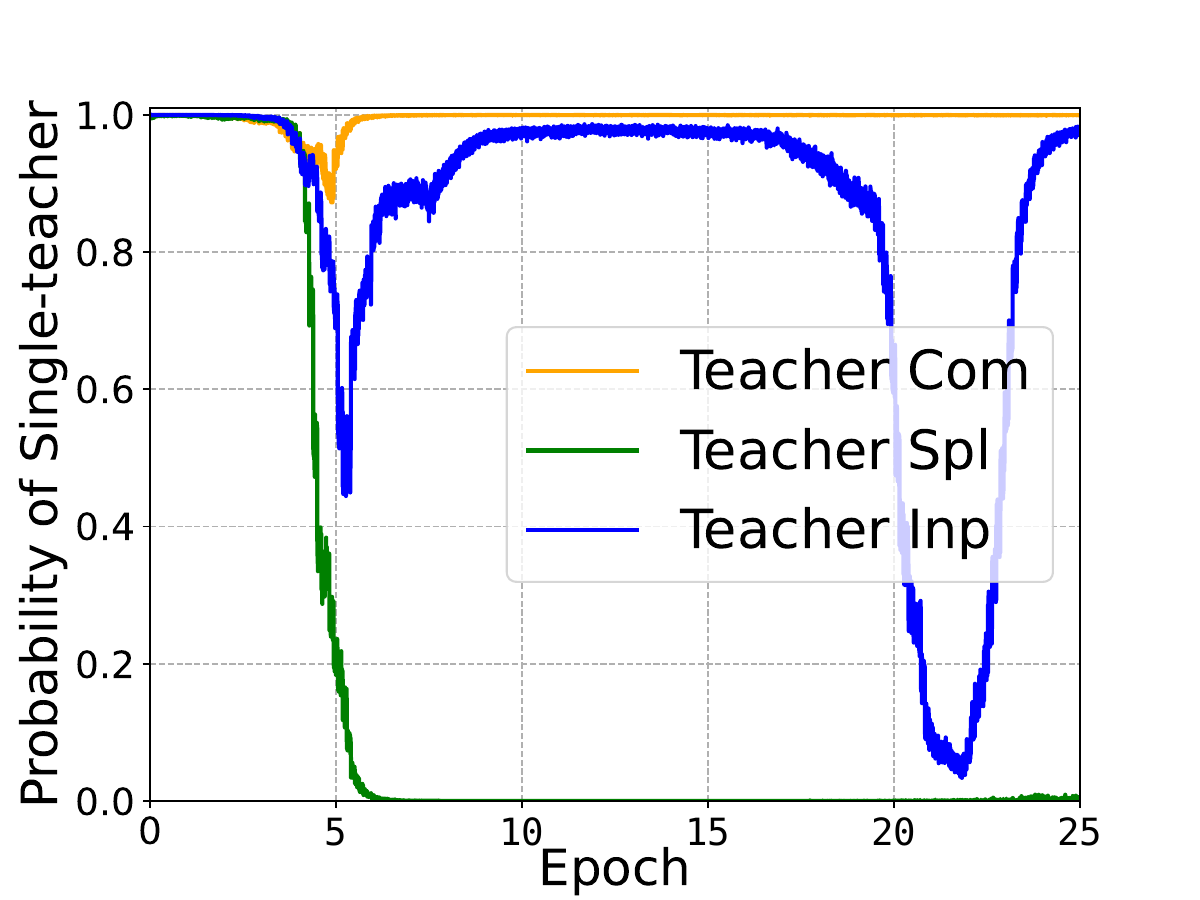}}
    \caption{Updates on teacher model's selection for Policy Network under different reward settings. Ours is also Setup \#9 in Tabel~4 of the main paper.}
    \label{fig:rl_pretraining}
\end{figure}

\noindent \textbf{Effectiveness of Specialized Teacher Models.}
As highlighted in the $(\cdot)$ in the upper part of Table~\ref{tab:single_teacher}, each specialized teacher model excels in detecting the corresponding type of tampered data.
Particularly noteworthy is the \textbf{Teacher-Com} model, which not only performs admirably on copy-move data but also demonstrates strong performance on splicing and inpainting data, leading to the best overall average performance. In contrast, the \textbf{Teacher-Inp} model exhibits minimal effectiveness on copy-move and splicing data, suggesting that inpainting data may exhibit a significant distributional divergence from other types of tampering data. Specifically, the processes of copy-move and splicing involve introducing altered content into the target image, whereas inpainting serves to remove such content.
As illustrated in the lower part of Table~\ref{tab:single_teacher}, the Re-MTKD framework has proven to enhance the performance of Cue-Net in capturing the commonalities and specifics of various tampering traces, when compared to the native Cue-Net. In particular, the performance improvement is significant in detecting inpainting data.
However, it is also evident that the teacher model sometimes does not successfully convey transfer knowledge to the student model, e.g., KD strategy Setup \#1 (\textbf{Single-Com}) in Table~4 of the main paper lags behind the F1 localization performance of \textbf{Teacher-Com} by 13.3\% on copy-move data.

\noindent \textbf{Dynamic Teacher Selection Analysis.}
To better understand the teacher selection process within the Re-DTS strategy, we visualize the selection probabilities of each teacher under different rewards in Setup \#9 and Setup \#5, as shown in Fig.~\ref{fig:rl_pretraining}.
In both settings, the distillation process tends to favor the selection of \textbf{Teacher-Com}, which aligns with its demonstrated ability to effectively handle multiple tampering traces, as outlined in Table~\ref{tab:single_teacher}.
For \textbf{Teacher-Spl}, both settings gradually decrease the level of guidance provided by the teacher after a certain number of update iterations. We hypothesize that \textbf{Teacher-Spl} and \textbf{Teacher-Com} have similar abilities in terms of detection range, but \textbf{Teacher-Com} may have a greater advantage.
Finally, for \textbf{Teacher-Inp}, Setup \#9 resulted in a more decisive reduction in the teacher's guidance during the update iteration. In contrast, Setup \#5 exhibited variability in its selection of teachers due to the absence of feedback on the teacher's knowledge transfer and the IFDL performance of the student model.
It is postulated that, despite the inpainting teacher's proficiency in this domain, the model gradually diminished the knowledge acquired from this teacher following a process of collective decision-making, due to the teacher's suboptimal performance in copy-move and splicing data.
\end{document}